\documentclass[11pt,a4paper]{article}
\usepackage[hyperref]{emnlp2020}
\usepackage{times}
\usepackage{latexsym}

\usepackage{wrapfig}
\usepackage{graphicx}
\usepackage{amsmath}
\usepackage{balance}
\usepackage{microtype}
\usepackage{subcaption}
\usepackage{tabularx}
\usepackage[shortlabels]{enumitem}
\usepackage[title]{appendix}
\usepackage{float}

\aclfinalcopy 

\title{Widget Captioning: Generating Natural Language Description for Mobile User Interface Elements}

\author{Yang Li\footnotemark[1], \space\space Gang Li\footnotemark[1], \space\space Luheng He,\space\space Jingjie Zheng,\space\space Hong Li\footnotemark[2],\space\space Zhiwei Guan\\
  Google Research, Mountain View, CA 94043, USA \\
  Georgia Tech, Atlanta, GA 30332, USA\footnotemark[2] \\
  \texttt{\{liyang, leebird, luheng, jingjiezheng, zguan\}@google.com}\\}
  
\date{}

\begin{document}
\maketitle

\interfootnotelinepenalty=10000
\makeatletter{\renewcommand*{\@makefnmark}{}
\footnotetext{$\ast$ Equal contribution}
\footnotetext{$\dagger$ Participated in the project during an internship at Google Research.}
}

\begin{abstract}
Natural language descriptions of user interface (UI) elements such as alternative text are crucial for accessibility and language-based interaction in general. Yet, these descriptions are constantly missing in mobile UIs. We propose widget captioning, a novel task for automatically generating language descriptions for UI elements from multimodal input including both the image and the structural representations of user interfaces. We collected a large-scale dataset for widget captioning with crowdsourcing. Our dataset contains 162,859 language phrases created by human workers for annotating 61,285 UI elements across 21,750 unique UI screens. We thoroughly analyze the dataset, and train and evaluate a set of deep model configurations to investigate how each feature modality as well as the choice of learning strategies impact the quality of predicted captions. The task formulation and the dataset as well as our benchmark models contribute a solid basis for this novel multimodal captioning task that connects language and user interfaces.
\end{abstract}

\section{Introduction}
Mobile apps come with a rich and diverse set of design styles, which are often more graphical and unconventional compared to traditional desktop applications. Language descriptions of user interface (UI) elements---that we refer to as \textit{widget captions}---are a precondition for many aspects of mobile UI usability. For example, accessibility services such as screen readers, e.g., \newcite{google_talkback},
rely on widget captions to make UI elements accessible to visually impaired users via text-to-speech technologies. Importantly, widget captions are an enabler for many language-based interaction capabilities such as voice commands and general screen understanding efforts.

However, a significant portion of mobile apps today lack widget captions in their user interfaces, which have stood out as a primary issue for mobile accessibility~\cite{Ross:2018:EIB:3234695.3236364,Ross:2017:EFL:3132525.3132547}. More than half of image-based elements have missing captions~\cite{Ross:2018:EIB:3234695.3236364}. Beyond image-based ones, our analysis of a UI corpus here showed that a wide range of elements  have missing captions. Existing tools for examining and fixing missing captions~\cite{apple_scanner,google_lint,Zhang:2018:RAM:3242587.3242616,Zhang:2017:IPR:3025453.3025846,Choo:2019:EAV:3290605.3300605} require developers to manually compose a language description for each element, which imposes a substantial overhead on developers. 

\begin{figure}
\centering
  \includegraphics[width=1.0\columnwidth]{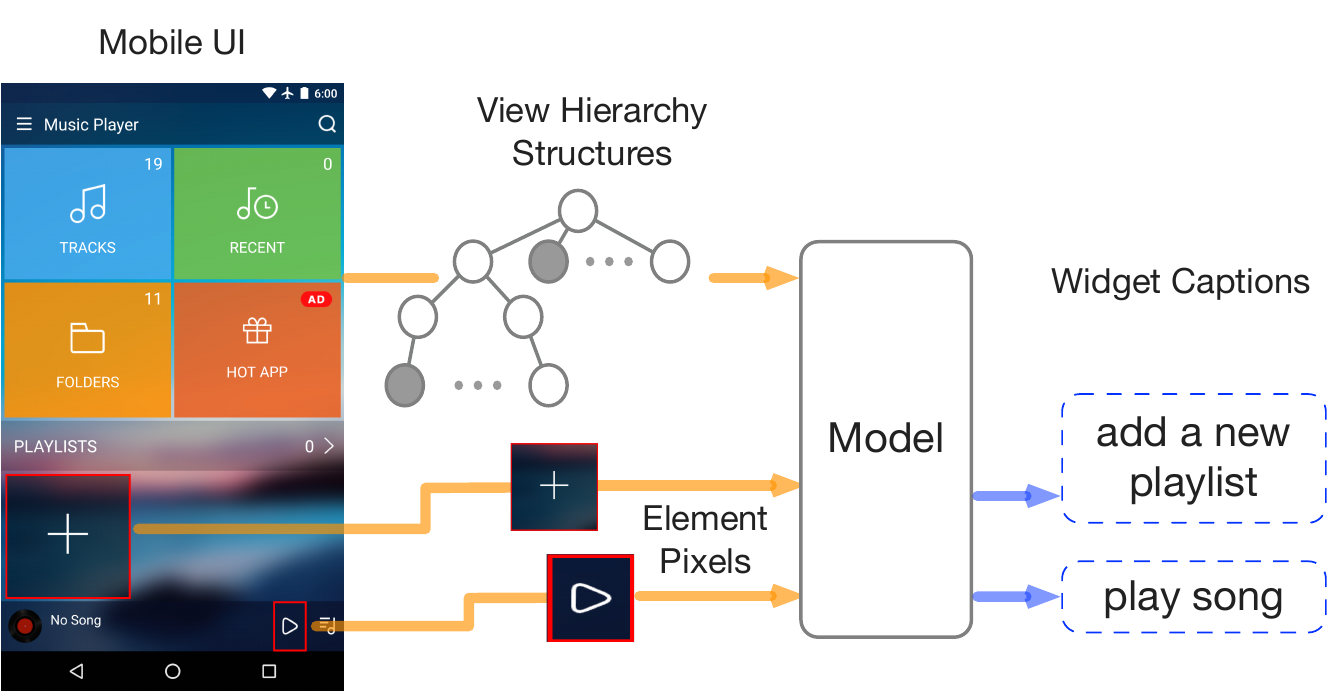}
  \caption{Widget captioning is a task to generate language descriptions for UI elements that miss captions, given multimodal input of UI structures and screenshot images. These captions are crucial for accessibility and language-based interaction in general.}~\label{fig:motivation}
\end{figure}
\setlength{\textfloatsep}{4pt}

We propose widget captioning, a novel task to automatically generate captions for UI elements\footnote{We use widgets and elements interchangeably.} based on their visual appearance, structural properties and context (see Figure \ref{fig:motivation}). This task is analogous to image captioning that generates language descriptions for images,
e.g., ~\citet{DBLP:journals/corr/XuBKCCSZB15,DBLP:journals/corr/LinMBHPRDZ14}. However, widget captioning raises several unique challenges. User interfaces are highly structural while traditional image captioning mostly focus on raw image pixels. Widget captioning is concerned with describing individual elements in the UI rather than the entire UI screen, while the entire screen provides useful contextual information for widget captioning. We target language generation for a broad set of UI elements, rather than only image-based ones. As we will show in our data analysis, many non-image elements also suffer the lack of captions. These challenges give rise to several interesting modeling questions such as how to combine both structural and image input and how to effectively represent each modality.

We start by processing and analyzing a mobile UI corpus. 
We then create a large dataset for widget captioning by asking crowd workers to annotate a collection of UI elements in the corpus.
Based on this dataset, we train and evaluate a set of model configurations to investigate how each feature modality and the choice of learning strategies would impact caption generation quality. 
Our champion model that is based on a Transformer~\cite{NIPS2017_7181} to encode structural information and a ResNet~\cite{he2015residual} for image input is able to produce accurate captions for UI elements based on both automatic and human evaluation. In summary, the paper makes the following contributions:

\begin{itemize}[nosep]
    \item We propose widget captioning as a task for automatically generating language descriptions for UI elements in mobile user interfaces; The task raises unique challenges for modeling and extends the popular image captioning task to the user interface domain.
    \item We create a dataset for widget captioning via crowdsourcing\footnote{Our dataset is released at https://github.com/google-research-datasets/widget-caption.}. It contains 162,859 captions created by human workers for 61,285 UI elements across 21,750 unique screens from 6,470 mobile apps. Our analysis on the missing captions and the linguistic attributes of collected captions contribute new knowledge for understanding the problem. 
    \item We investigate a set of model configurations and learning strategies for widget captioning; our benchmark models leverage multimodal input including both structural information and images of user interfaces\footnote{Our model code is released at https://github.com/google-research/google-research/tree/master/widget-caption.}. They are able to generate accurate captions for UI elements, and yet leave enough room for improvement for future research.
\end{itemize}

\section{Related Work}
The lack of captions or alternative text has been a universal problem in user interfaces, ranging from mobile apps to web pages~\cite{Ross:2018:EIB:3234695.3236364,Ross:2017:EFL:3132525.3132547,Gleason:2019:LAL:3308558.3313605,Guinness:2018:CCE:3173574.3174092}. Based on an analysis of an Android UI corpus~\cite{Deka:2017:RMA:3126594.3126651}, \newcite{Ross:2018:EIB:3234695.3236364} revealed that a significant portion of image-based buttons lack alternative text. By examining a broader set of UI elements, we found missing captions is a general issue across UI element categories. 



Automatic image captioning has been a classic task where a model learns to generate language descriptions for images, which has gained substantial progress with the advance of deep learning~\cite{10.5555/2566972.2566993,7558228,DBLP:journals/corr/KarpathyF14,DBLP:journals/corr/XuBKCCSZB15,DBLP:journals/corr/LinMBHPRDZ14}, and the availability
of datasets such as Flickr30K~\cite{young-etal-2014-image} and MS-COCO~\cite{DBLP:journals/corr/LinMBHPRDZ14}.
In contrast to image captioning, widget captioning that we propose is concerned with describing individual elements in the context of the UI screen. In addition to image input, 
widget captioning has access to UI structures such as view hierarchies. These raise unique modeling opportunities for multimodal captioning. 

Many image captioning models~\cite{DBLP:journals/corr/XuBKCCSZB15,conceptual_captioning} involve an encoding net and a language generation net. The encoding net is typically a deep Convolutional Neural Network (CNN)~\cite{NIPS2012_4824} that encodes the image input as a collection of latent vector representations. The generation net is often an auto-regressive decoding model, enhanced with neural attention. For widget captioning, the encoding net needs to encode multimodal input that include both images and UI structures. For UI structure encoding, 
recent work~\cite{seq2act} investigated both Graph Convolutional Network (GCN)~\cite{pmlr-v48-niepert16} and Transformer~\cite{NIPS2017_7181} and showed that a Transformer encoder gives better performance on the task, which we will use in this work. 

Our learning strategy is akin to BERT~\cite{devlin-etal-2019-bert} that uses the words in the same sentence to predict those missing using Transformer, to learn a contextual word representation. In our case, we use the information of elements in the same screen context to predict those with missing captions. To generate captions, based on the encoder output, we run multiple instances of the decoding model in parallel, one for each element to be captioned.


 

\section{Creating the Widget Caption Dataset}
We first create a mobile UI corpus, 
and then ask crowd workers to create captions for UI elements that have missing captions, which is followed by a thorough analysis of the dataset.

\subsection{Creating a Mobile UI Corpus}
We create a mobile UI corpus based on RICO, a public dataset of Android user interfaces, which has 66K screens collected from human users interacting with Android devices~\cite{Deka:2017:RMA:3126594.3126651}, which include top apps selected broadly from various categories in Google Play Store. We expanded the dataset using a crawling robot to perform random clicks on mobile interfaces, which added 12K novel screens to our corpus. Each screen comes with both a screenshot JPG/PNG image and a view hierarchy\footnote{\url{https://developer.android.com/reference/android/view/View}} in JSON. The view hierarchy is a structural tree representation of the UI where each node has a set of properties such as content description, class information, visibility, and bounding boxes. 

\subsubsection{Preprocessing the UI Corpus}
We first exclude UI screens with missing or inaccurate view hierarchies, which could occur when Android logging is out of sync. This filtering step was conducted by asking crowd workers to visually examine each UI and confirm that the bounding boxes of all the leaf nodes in the hierarchy match the UI elements shown on the screenshot image. We focus on leaf nodes because most interactive elements are leaf nodes. The filtering process resulted in 24,571 unique screens from 6,853 mobile apps.

We then select UI elements that are visible and clickable because they are responsible for many of the interaction tasks. Similar to previous work, we consider an element missing captions when both its \texttt{contentDescription} and \texttt{text} properties in the view hierarchy are missing, according to the Android accessibility guideline\footnote{\url{https://developer.android.com/guide/topics/ui/accessibility/apps}}. Screen readers such as the TalkBack service$^1$ rely on these fields to announce the widget. 
Overall, in our dataset, there are 74,379 UI elements with missing captions, across 10 categories of UI elements (see Figure \ref{fig:label_ratio}). 

\subsubsection{Understanding Missing Captions}
Previous work analyzed missing captions for image-based elements~\cite{Ross:2018:EIB:3234695.3236364}. We include all types of elements in our dataset and analysis (see Appendix~\ref{app:proc_corpus}). The results from analyzing image-based elements in our corpus are comparable to previous analysis, i.e., 95\% of Floating Action Buttons, 83\% of Image Views, and 57\% of Image Buttons have missing captions. Beyond these image-based elements, we found that missing captions is a serious issue for other types of elements as well (see Figure \ref{fig:label_ratio}). More than 50\% of the \texttt{Switch}, \texttt{Compound Button}, \texttt{Check Box} and \texttt{Toggle Button} have missing captions. 24.3\% of the screens have none pre-existing captions.

\subsection{Crowdsourcing Widget Captions}
To best match the target scenario of predicting for elements with missing captions, we asked crowd workers to created captions for these elements, which are used as labels for training and testing. Because pre-existing captions in the corpus are not always correct, they are used as model input, to provide the context, but not as output. 

We developed a web interface for crowd workers to create language descriptions for UI elements that have missing captions. The interface shows a screenshot of the mobile interface, with the UI element that needs to be captioned highlighted (see Appendix~\ref{app:annotation_ui}). Workers can input the caption using a text field, or indicate that they cannot describe the element. In the annotation guidelines, we asked the workers to caption the element for vision-impaired users to understand its functionalities and purposes. The captions need to be concise but more descriptive than generic words such as ``button" or ``image". We recruited over 5,454 workers from Amazon Mechanical Turk\footnote{mturk.com} 
over multiple batches. While the elements to be labeled by each worker are randomly selected, we instrumented the task in the way such that a worker can only label each unique element once, and each element is labeled by 3 different workers.

\subsection{Data Analyses}

\begin{figure}[t]
\centering
  \includegraphics[width=\columnwidth]{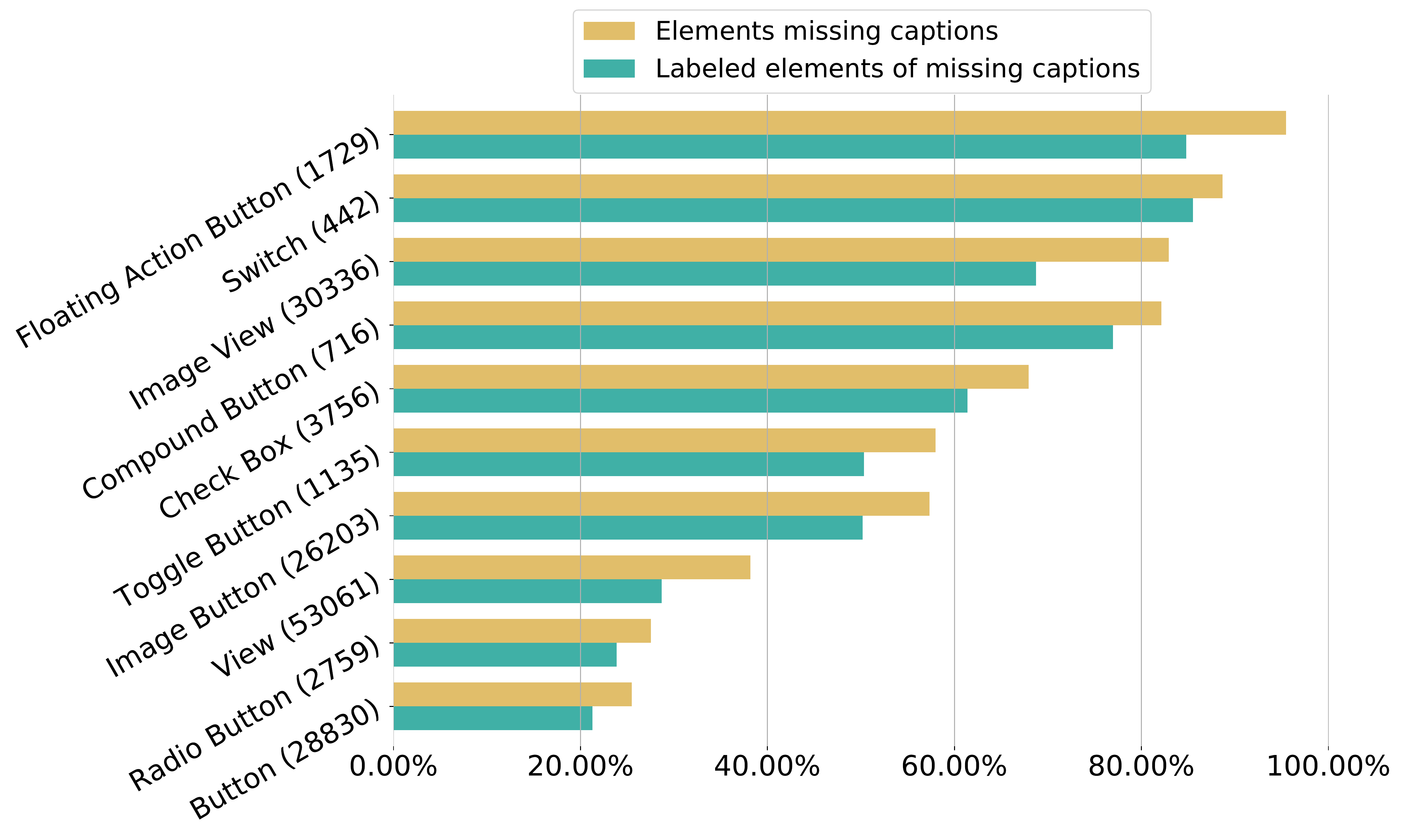}
  \caption{The percentage of elements that have missing captions (red) for each category and elements labeled by MTurk workers (green). The numbers in parentheses are total counts of the elements.}~\label{fig:label_ratio}
\end{figure}

Human workers can skip elements when they were not sure how to describe them. For all the elements of each type given to workers, the percentage of elements being captioned  ranges from 75\% to 94\% (see Figure~\ref{fig:label_ratio}). In particular, the \texttt{View} type has the lowest labeling ratio of 75\%, which we suspect that elements with the \texttt{View} type, a generic widget type, tend to be quite arbitrary and are difficult for the workers to understand. We only kept the elements that received at least 2 captions (from different workers). On average, each element received 2.66 captions. In total, we collected 162,859 captions for 61,285 UI elements across 21,750 unique screens, from 6,470 mobile apps. 

\begin{figure}[H]
\centering
  \includegraphics[width=0.95\columnwidth]{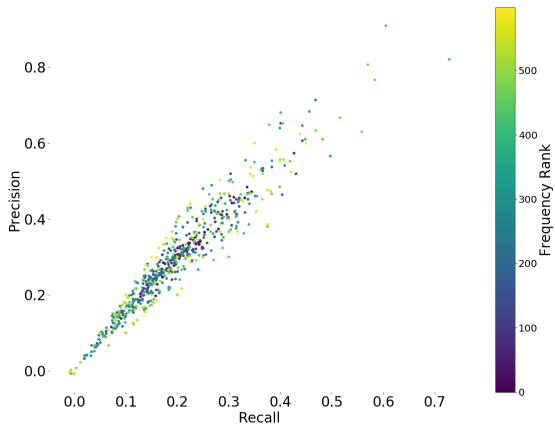}
  \caption{The distribution of precision and recall for the top 6K words of the collected captions.}~\label{fig:agreement}
\end{figure}

To measure inter-annotator agreement, we computed the word-level precision and recall for all the words with two or more occurrences in the collected captions (see Figure \ref{fig:agreement}), as in the COCO image captioning dataset ~\cite{Chen2015-cy}. The results were generated on about 6K words, which amount to 98.6\% of all the word occurrences in the captions. Figure \ref{fig:agreement} shows that our corpus has reasonable word-level agreement among the captions of the same widget.  Specifically, for the 6K words, we report the mean precision and recall of every 10 consecutive words in the vocabulary. Therefore, we have 600 data points, each representing precision/recall of 10 words. The ranks of the words in the vocabulary sorted by word frequency are used to color the data points. Lower rank indicates higher word frequencies in the corpus.


\subsubsection{Caption Phrase Analysis}
We analyzed the distribution of caption lengths created by human workers (see Figure~\ref{fig:length_distribution}). We found most captions are brief, i.e., two to three words. But a significant number of captions have more words, which are often long-tail captions. 
The average length of captions from human workers is 2.72. Overall, the length distribution of captions created by human workers is similar to those preexisting in the UI corpus, which are from app developers (see Appendix~\ref{app:phrase_dist}). The latter will be used as a feature input to the model, which we will discuss later.

The captions in our dataset include a diverse set of phrases. The most frequent caption is ``go back'' that amounts to 4.0\% of the distribution. Other popular captions among the top 5 are ``advertisement'' (2.4\%), ``go to previous'' (0.8\%), ``search'' (0.7\%) and ``enter password'' (0.6\%). 

A common pattern of the phrases we observe is \textit{Predicate + Object}. Table \ref{tab:phrase_type} lists the 7 common predicates and their most frequent objects. As we can see, the phrases describe highly diverse functionalities of the UI elements. It is difficult to classify them into a few common categories. This linguistic characteristics motivated us to choose sequence decoding for caption generation instead of classification based on a predefined phrase vocabulary. The diversity of caption phrases indicates that widget captioning is a challenging machine learning task.

\begin{figure}
\centering
  \includegraphics[width=0.8\columnwidth]{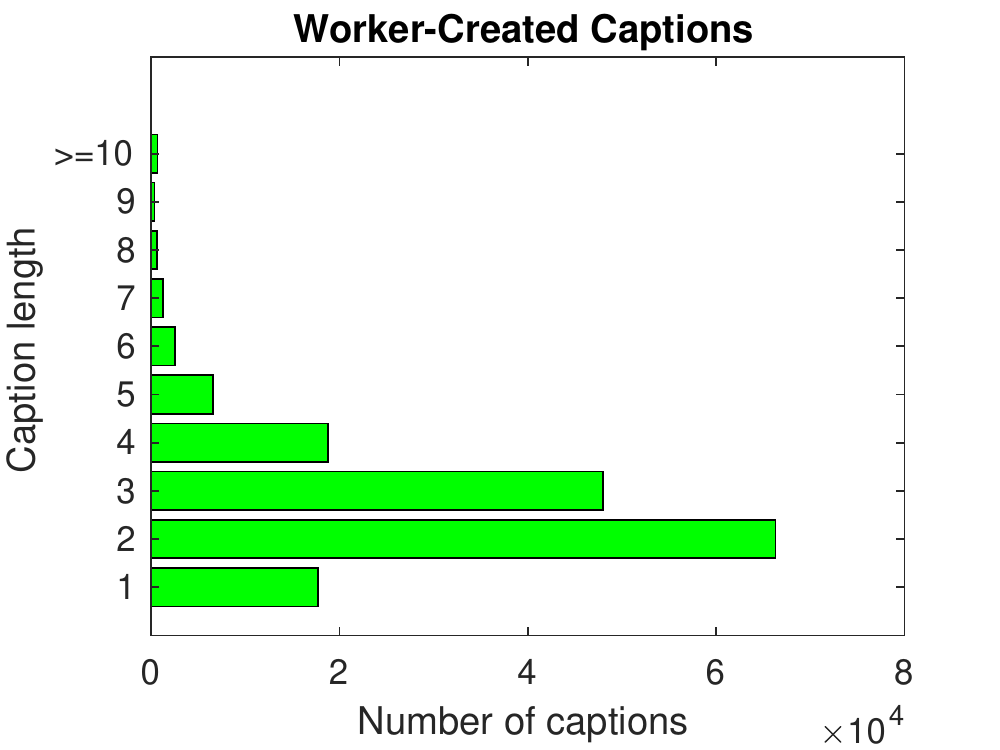}
  \caption{The length distribution of captions created by human workers. The X axis shows the number of captions and the Y axis is the lengths from $1$ to $\geq{10}$.}~\label{fig:length_distribution}
\end{figure}

\begin{table*}[h]
  \centering
  \small
  \def\arraystretch{1.2}
  \begin{tabular}{l|l}
    \hline
    {\textbf{Predicate}} & {\textbf{Object}} \\
    \hline
   search & location, contact, app, music, map, image, people, recipe, flight, hotel \\
   enter & password, email, username, phone, last name, first name, zip code, location, city, birthday \\
   select & image, ad, color, emoji, app, language, folder, location, ringtone, theme \\
   toggle & autoplay, favorite, menu, sound, advertisement, power, notification, alarm, microphone, vibration \\
   share (to) & article, facebook, twitter, image, app, video, instragram, recipe, location, whatsapp\\
   download & app, sound, song, file, image, video, theme, game, wallpaper, effect \\
   close & window, ad, screen, tab, menu, pop-up, notification, file, settings, message \\
        \hline

  \end{tabular}
  \caption{In our dataset, the popular predicates are often associated with a diverse set of objects that are contextually determined.
  }~\label{tab:phrase_type}
  \vspace{-2pt}
\end{table*}

\begin{figure}[h]
\centering
  \begin{subfigure}[b]{0.5\columnwidth}
  \includegraphics[width=0.95\columnwidth]{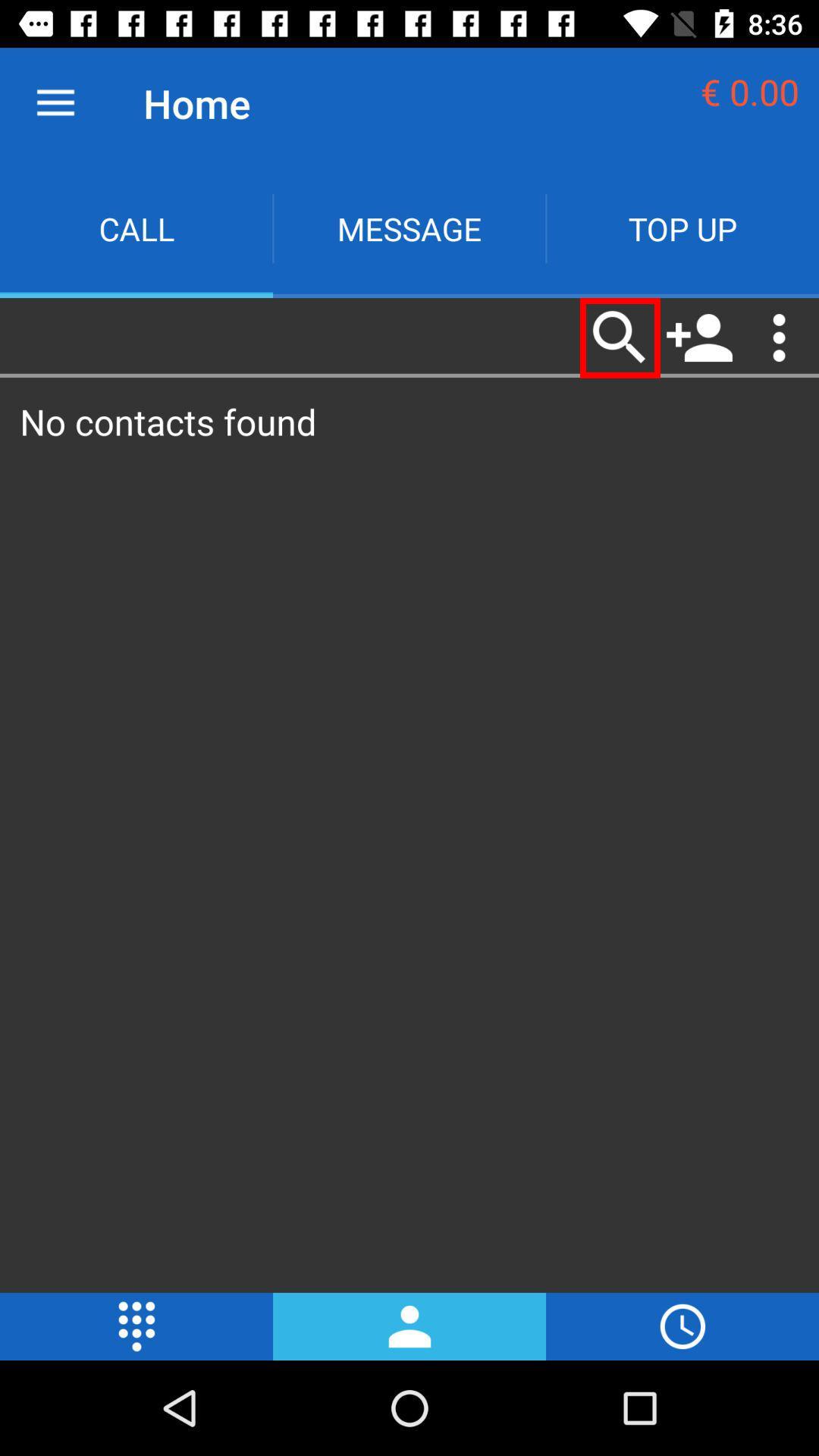}
  \end{subfigure}%
\begin{subfigure}[b]{0.5\columnwidth}
  \includegraphics[width=0.95\columnwidth]{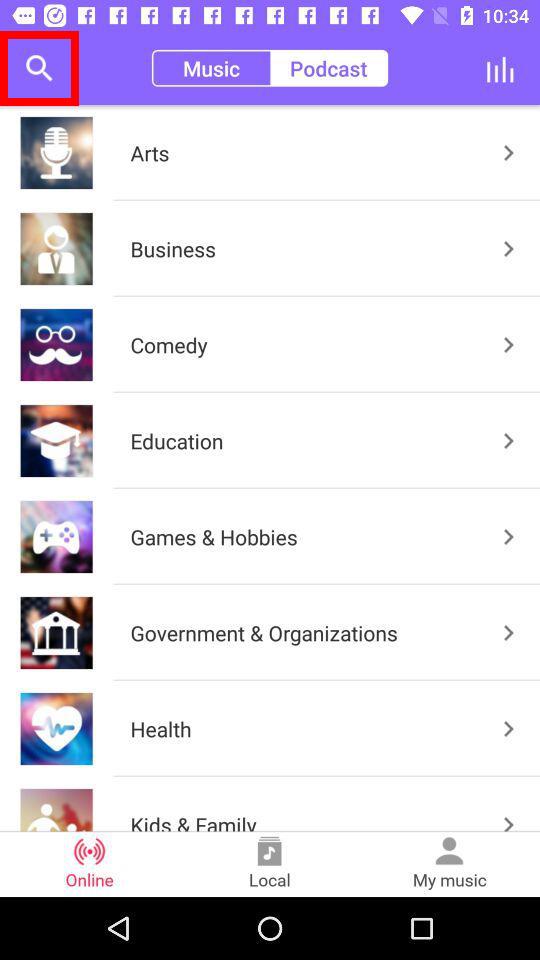}
  \end{subfigure}%
  
\caption{Two UI elements (outlined in red) of ``search" predicate. Left: search contact; Right: search music.}~\label{fig:search_predicate}

\end{figure}

Furthermore, these examples in Table~\ref{tab:phrase_type} show that, to distinguish different objects for the same predicate, it is necessary to take into account the screen context that the element belongs to. For example, Figure~\ref{fig:search_predicate} shows two examples of the ``search" predicate. The two UI elements have very similar images (magnifiers) although they are for searching different objects. Thus context information is critical for models to decode the correct objects.

\vspace{-5pt}
\subsubsection{View Hierarchy Complexities}
\begin{figure}[h]
\centering
  \includegraphics[width=0.5\columnwidth]{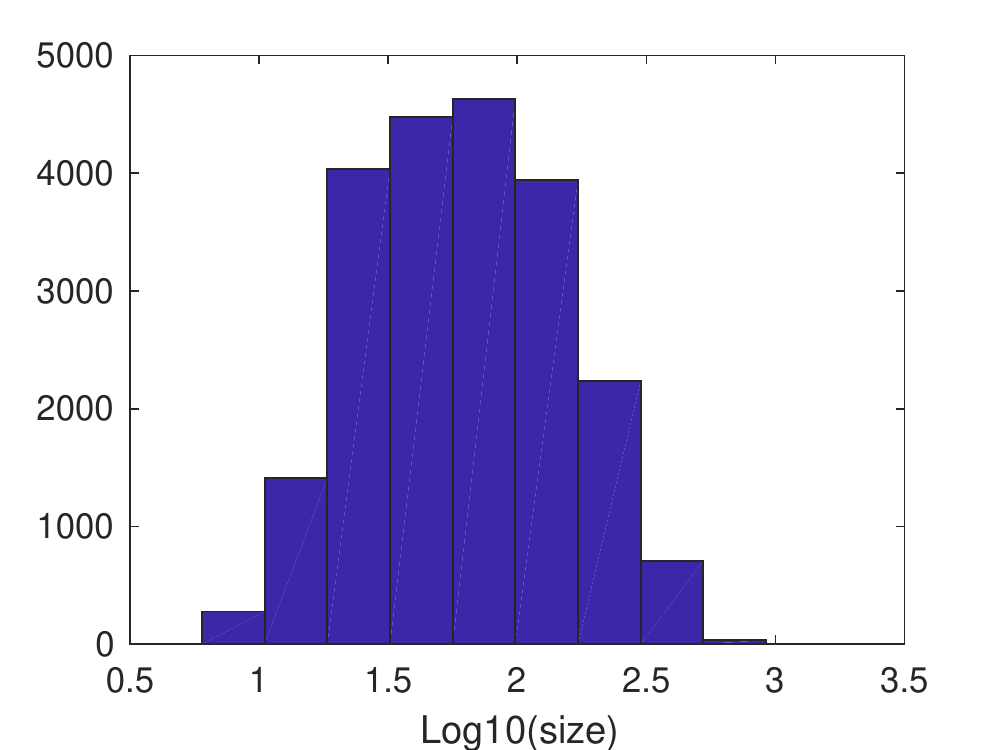}\includegraphics[width=0.5\columnwidth]{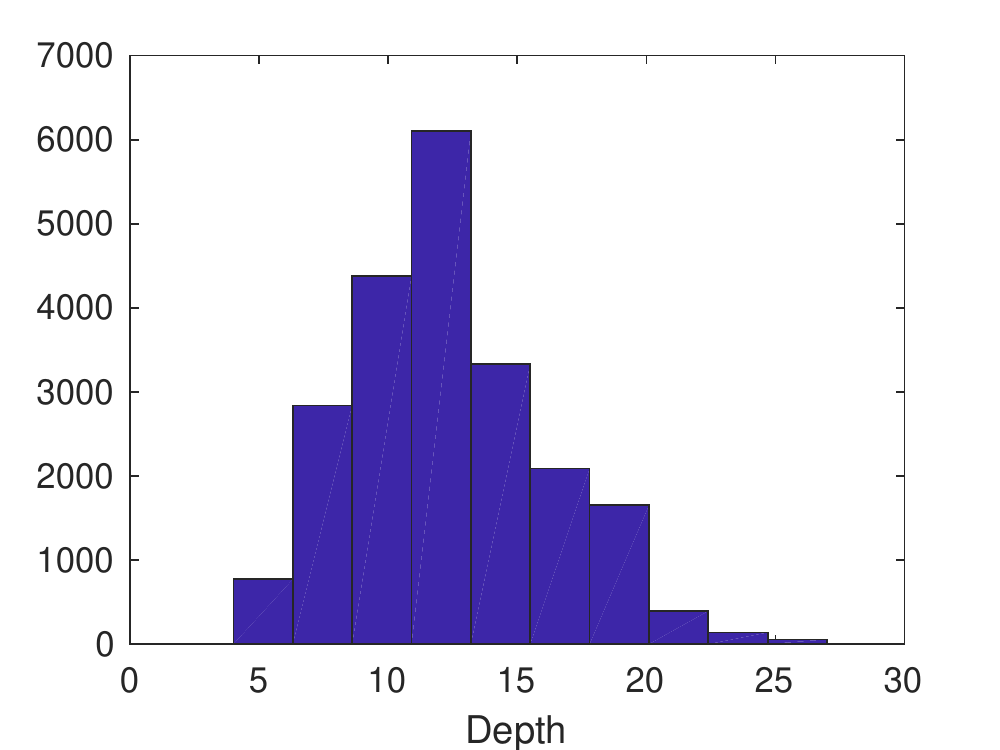}
  \caption{The histogram of $\log_{10}$ transformed view hierarchy sizes on the left, and the histogram of tree depths on the right.}~\label{fig:graphsize}
\end{figure}

A unique modality in widget captioning is UI structures as represented by view hierarchy trees. To better understand the complexity of the UI structures, we analyze the size and depth of the view hierarchy of each UI. 
The size of a view hierarchy is the total number of nodes in the hierarchy tree, including both non-terminal nodes, i.e., layout containers, and leaf nodes. The size distribution is highly skewed and with a long tail towards large view hierarchies (see the left of Figure~\ref{fig:graphsize}). The median size of view hierarchies is 61, with a minimum of 6 and a maximum of 1,608 nodes.
Many view hierarchies have a large depth with a median depth of 11, a minimum of 3 and a maximum of 26 (see the right of Figure~\ref{fig:graphsize}). These show that view hierarchies are complex and contain rich structural information about a user interface.

\begin{figure}[!htb]
  \includegraphics[width=0.95\columnwidth]{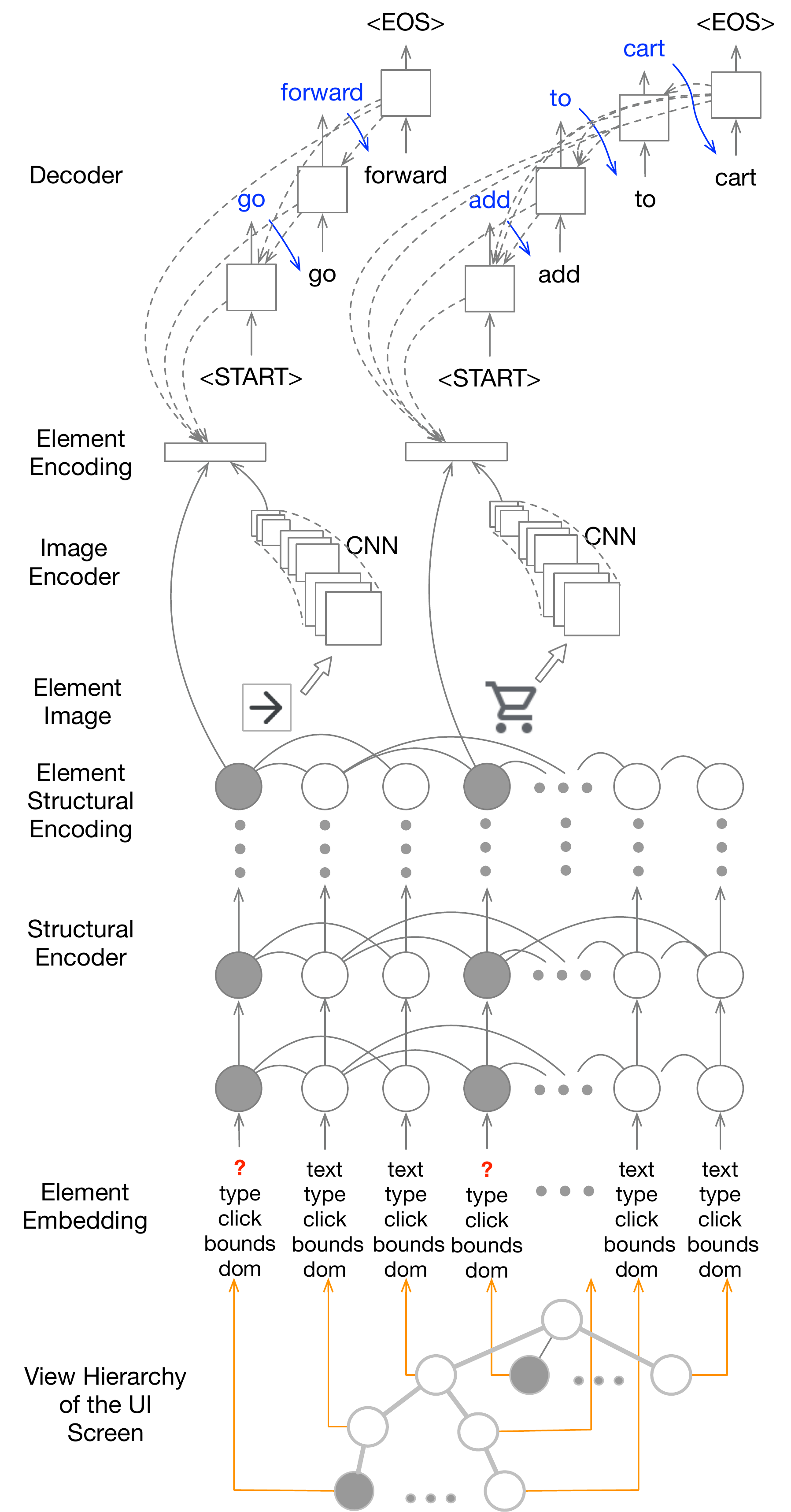}
  \caption{Our widget captioning model takes both view hierarchy structures and element image input, and performs parallel decoding for multiple elements on the screen missing captions. The shaded nodes represent the elements missing captions.}~\label{fig:architecture}
\end{figure}

\section{Widget Captioning Models}
To understand the challenges and feasibility of automatic widget captioning, we investigate deep models for this task. Captioning models are often designed based on an encoder-decoder architecture. We formulate widget captioning as a multimodal captioning task where the encoder takes both the structural information and the pixel appearance of the UI element, and the decoder outputs the caption based on the encodings (see Figure~\ref{fig:architecture}).


\subsection{Encoding the Structural Information}

We hypothesize that the relationship of UI elements on the screen provides useful contextual information for representing each object thus benefits captioning. We use a Transformer model~\cite{NIPS2017_7181} to encode the set of elements on a screen, which learns how the representation of an element should be affected by the others on the screen using multi-head neural attention. The input to a Transformer model requires both the content embedding and positional encoding. Similar to previous work~\cite{seq2act}, we derive these embeddings for each element on the screen in the following manner. 

Each UI element in the view hierarchy consists of a tuple of properties. The \texttt{widget\_text} property includes a collection of words possessed by the element.
We acquire the embedding of the \texttt{widget\_text} property of the $i$-th element on the screen, $e_i^{X}$, by max pooling over the embedding vector of each word in the property. When the \texttt{widget\_text} property is empty, i.e., the element is missing a caption, a special embedding, $e^{\emptyset}$, is used. With $e_{i}^T$, the embedding of the \texttt{widget\_type} property (see Figure~\ref{fig:label_ratio}), and $e_{i}^C$, the embedding of whether the widget is clickable, $[e_{i}^X; e_{i}^T; e_{i}^C]$ form the content embedding of the element.

The \texttt{widget\_bounds} property contains four coordinate values on the screen: \texttt{left}, \texttt{top}, \texttt{right} and \texttt{bottom}, which are normalized to the range of $[0, 100)$. The \texttt{widget\_dom} property contains three values describing the element tree position in the view hierarchy: the sequence position in the \texttt{preorder} and the \texttt{postorder} traversal, and the  \texttt{depth} in the view hierarchy tree. These are all treated as categorical values and represented as embedding vectors. The sum of these coordinate embeddings form the positional embedding vector of the element, $e_i^{B}$. 

The concatenation of all these embeddings forms the representation of a UI element: $e_i=[e_i^{X}; e_i^{T}; e_{i}^C; e_i^{B}]W^E$, where $W^E$ is the parameters to linearly project the concatenation to the dimension expected by the Transformer model. The output of the Transformer encoder model, $h_{i}$, is the structural encoding of the $i$-th element on the screen.

\subsection{Encoding Element Images}

The image of an element is cropped from the UI screenshot and rescaled to a fixed dimension, which results in a \texttt{64x64x1} tensor, where \texttt{64x64} are the spatial dimensions and \texttt{1} is the grayscale color channel. The image dimension strikes a good balance for representing both small and large elements, which preserves enough details for large elements after scaled down and enables a memory footprint good for model training and serving.

We use a ResNet (CNN)~\cite{he2015residual} to encode an element image.
Each layer in the image encoder consists of a block of three sub layers with a residual connection---the input of the 1st sub layer is added to the input of the 3rd sub layer. There are no pooling used, and instead, the last sub layer of each block uses stride 2 that halves both the vertical and horizontal spatial dimensions after each layer. At the same, each layer doubles the channel dimension, starting from the channel dimension 4 of the first layer. Most sub layers use a kernel size of $3\times{3}$ except the initial and ending sub layers in the first layer that use a kernel size of $5\times{5}$. We will discuss further details of model configuration for the image encoder in the experiment section. The output of the multi-layer CNN is the encoding vector of the element image, which we refer to as $g_{i}$ for the $i$-th element.

\subsection{Decoding Captions}
We form the latent representation of the $i$th element on the screen by combining its structural and image encoding: $z_{i}=\sigma([h_{i}^{L};g_{i}],\theta^{z})W^z$, where $\sigma(\cdot)$ is the non-linear activation function parameterized by $\theta^{z}$ and $W^z$ is the trainable weights for linear projection. Based on the encoding, we use a Transformer~\cite{NIPS2017_7181} decoder model for generating a varying-length caption for the element. 
\vspace{-5pt}
\begin{align*}
    a_{i,1:M}^{l}&=\mbox{Masked\_ATTN}(x_{i,1:M}^{l},W_{d}^{Q},W_{d}^{K},W_{d}^{V}) \\
    x_{i,1:M}^{l+1} &=\mbox{FFN}(a_{i,1:M}^{l}+z_{i}, \theta_{d})
\end{align*}

\noindent where $0\leq{l}\leq{L}$ is the layer index and $M$ is the number of word tokens to decode. $x_{i,1:M}^{0}$, the input to the decoder model, is the token embedding with the sequence positional encoding. $W_{d}^{Q}$, $W_{d}^{K}$, and $W_{d}^{V}$ are trainable parameters for computing the queries, keys, and values. Masked\_ATTN in a Transformer decoder allows multi-head attention to only attend to previous token representations. The element encoding, $z_i$, is added to the attention output of each decoding step, $a_{i,1:M}^{l}$, before feeding into the position-wise, multi-layer perception (FFN), parameterized by $\theta_{d}$. The probability distribution of each token of the caption is finally computed using the softmax over the output of the last Transformer layer: $y_{i,1:M}=\mbox{softmax}(x_{i,1:M}^{L}W_d^{y})$ where $W_d^{y}$ is trainable parameters.

There is one instance of the decoder model for each element to be captioned.  The captions for all the elements with missing captions on the same screen are decoded in parallel. The entire model, including both the encoder and decoder, is trained end-to-end, by minimizing $\mathcal{L}_{screen}$, the average cross entropy loss for decoding each token of each element caption over the same screen.
\begin{equation*}
    \mathcal{L}_{screen}=\frac{1}{|\nabla|}\sum_{i\in{\nabla}}\frac{1}{M}\sum_{j=1}^{M}\mbox{Cross\_Entropy}(y_{i,j}^{'}, y_{i,j})
\end{equation*}
\noindent where $\nabla$ is the set of elements on the same screen with missing captions and $y_{i,j}^{'}$ is the groundtruth token. Training is conducted in a teacher-forcing manner where the groundtruth caption words are fed into the decoder. During prediction time, the model decodes autoregressively.

\section{Experiments}
We first discuss the experimental setup, and then report the accuracy of our model as well as an analysis of the model behavior.

\subsection{Datasets}

We split our dataset into training, validation and test set for model development and evaluation, as shown in Table \ref{tab:dataset}. The UIs of the same app may have a similar style. To avoid information leaks, the split was done app-wise so that all the screens from the same app will not be shared across different splits.  Consequently, all the apps and screens in the test dataset are unseen during training, which allow us to examine how each model configuration generalizes to unseen conditions at test. 

Our vocabulary includes 10,000 most frequent words (that covers more than 95\% of the words in the dataset), and the rest of the words encountered in the training dataset is assigned a special unknown token \texttt{<UNK>}. During validation and testing, any \texttt{<UNK>} in the decoded phrase is removed before evaluation. Since each element has more than one caption, one of its captions is randomly sampled each time during training. For testing, all the captions of an element constitute its reference set for computing automatic metrics such as CIDEr.

\begin{table}[h]
\small
\def\arraystretch{1.3}
  \centering
  \begin{tabular}{l|r|r|r|r}
    \hline
    {Split} & {Apps} & {Screens} & {Widgets} & {Captions} \\
    \hline
   Training &  5,170 & 18,394 & 52,178 & 138,342 \\
   Validation & 650 & 1,720 & 4,548 & 12,242 \\
   Test & 650 & 1,636 & 4,559 & 12,275 \\
   \hline
   Total & 6,470 & 21,750 & 61,285 & 162,859 \\
    \hline
  \end{tabular}
  \caption{Dataset statistics.}~\label{tab:dataset}
  \vspace{-5pt}
\end{table}

The training, validation and test datasets have a similar ratio of 40\% for caption coverage, i.e., the number of elements with preexisting captions with respect to the total number of elements on each screen, with no statistical significance ($p>0.05$). Screens with none preexisting captions exist in all the splits.




\subsection{Model Configurations}
We based our experiments on Transformer as it outperformed alternatives such as GCNs and LSTMs in our early exploration. We tuned our model architectures based on the training and validation datasets. We initialize the word embeddings with pre-trained 400K-vocab 300-dimensional GLOVE embeddings~\cite{pennington2014glove}, which are then projected onto a 128-dimensional vector space. To reduce the number of parameters needed in the model, the embedding weights are shared by both the structural encoder and the decoder.
Both the Transformer structural encoder and the Transformer decoder use 6 Transformer layers with a hidden size of 128 and 8-head attention. We used a 7-layer ResNet for encoding the pixels of a target element, where each layer consists of 3 sub layers as discussed earlier, which in total involves 21 convolutional layers and the output of the final layer is flattened into a 256-sized vector. We used batch normalization for each convolutional layer.
The final encoding $z_{i}$ of an element is a 128-dimensional vector that is used for decoding. See Appendix~\ref{app:config_train} for training details.

\subsection{Metrics \& Results}
We report our accuracy based on BLEU (unigram and bigram) ~\cite{Papineni2002-bw}, CIDEr ~\cite{DBLP:journals/corr/VedantamZP14a}, ROUGE-L ~\cite{Lin:2004:OME:1220355.1220427} METOER ~\cite{Denkowski2014-mu} and SPICE~\cite{DBLP:journals/corr/AndersonFJG16} metrics (see Table~\ref{tab:accuracy}). For all these metrics, a higher number means better captioning accuracy---the closer distances between the predicted and the groundtruth captions.

We investigate how model variations impact the overall accuracy of captioning (Table \ref{tab:accuracy}). 
\textit{Template Matching} is an obvious baseline, which predict the caption of an element based on its image similarity with elements that come with a caption. We use pixel-wise cosine similarity to compare the element images. Although this heuristic-based method is able to predict captions for certain elements, it performs poorly compared to the rest of the models that use deep architectures. \textit{Pixel Only} model, which only uses the image encoding of an element, performs significantly better than \textit{Template Matching}, which indicates that image encoding, $g_{i}$, is a much more efficient representation than raw pixels.

\textit{Pixel+Local}, which uses both image encoding, $g_{i}$, and the structural representation computed only based on the properties of the element, offers further improvement on the accuracy. 
Our full model, \textit{Pixel+Local+Context}, uses both image encoding, $g_{i}$, and the screen context encoding, $h_i$. It achieves the best results, which indicate that screen context carries useful information about an element for generating its caption. Among all the structural features, the \texttt{widget\_text} property plays an important role (see the ablation study in Appendix~\ref{app:ablation}). 

\begin{table*}[h]
\small
\centering
\begin{tabularx}{0.82 \textwidth}{l|c|c|c|c|c|c}
    \hline
    {Model Configuration}
    & {BLEU-1}
    & {BLEU-2}
    & {ROUGE}
    & {CIDEr}
    & {METOER}
    & {SPICE} \\
    \hline
   & \multicolumn{6}{c}{Full Test Set} \\
    \hline

    Template Matching & 20.2 & 11.2 & 20.9 & 38.0 & 13.2 & 6.5 \\
    Pixel Only & 35.6 & 24.6 & 35.6 & 71.3 & 24.9 & 11.2 \\
    Pixel+Local & 42.6 & 29.4 & 42.0 & 87.3 & 29.4 & 15.3 \\
    Pixel+Local+Context (PLC) & \textbf{44.9} & \textbf{32.2} & \textbf{44.7} & \textbf{97.0} & \textbf{31.7} & \textbf{17.6}  \\
    PLC Classification & 36.2 & 25.7 & 36.9 & 78.9 & 26.0 & 13.6 \\
   \hline
   & \multicolumn{6}{c}{Predicate-Object Subset} \\
   \hline
Template Matching & 20.8 & 11.2 & 21.3 & 34.5 & 12.6 & 7.5 \\
Pixel Only & 39.4 & 27.2 & 39.1 & 69.6 & 25.8 & 14.2 \\
Pixel+Local & 48.5 & 34.8 & 47.4 & 94.7 & 32.3 & 19.9 \\
Pixel+Local+Context (PLC) & \textbf{52.0} & \textbf{38.8} & \textbf{51.3} & \textbf{110.1} & \textbf{36.4} & \textbf{23.3}  \\
PLC Classification & 38.5 & 27.0 & 38.4 & 78.9 & 26.3 & 16.8 \\
  \hline
\end{tabularx}
  \caption{The accuracy of each model configuration on the full set and the predicate-object subset of the test dataset.} 
  \label{tab:accuracy}
\end{table*}


In addition to examining the impact of input modality on captioning quality, we compare strategies of caption generation: sequence decoding based on word tokens versus classification based on common caption phrases. \textit{PLC Classification} model uses the same input modality and encoding as \textit{Pixel+Local+Context} but decodes a single predefined phrase based on a vocabulary of top 10K caption phrases---the same size as the token vocabulary for decoding. It performed poorly compared to the decoding-based approach.

To further validate the usefulness of the context and the information from view hierarchy, we evaluate the models on a subset of UI elements with one of their reference caption phrases is of the \textit{Predicate + Object} pattern (see Table \ref{tab:phrase_type}). This subset consists of about 40\% of the UI elements from the test set (see Appendix~\ref{app:predicate_obj_phrase} for details). All the models achieve better accuracy because the predicate-object subset consists of more common words. \textit{Pixel+Local+Context} remains the champion model, and more importantly, acquired the most significant gain across all the metrics (see Table~\ref{tab:accuracy}). This indicates that context information plays a crucial role for generating this type of captions whose object parts need to be contextually determined. In contrast, \textit{PLC Classification} still performs worse than the champion decoding-based model. While the subset contains more common words, their combination can form long-tail phrases. A classification-based method such as \textit{PLC Classification} is more vulnerable to the data sparsity of long-tail phrases. 

\subsection{Human Evaluation}

To assess the quality of the generated phrases by human, we asked another group of crowd workers to manually verify the model generated captions for the entire test set, by presenting each human rater a caption and its corresponding element in a UI screenshot. For each phrase, we asked three raters to verify whether the caption phrase correctly describes the functionality and purpose of the element. We compared two of our models and the results are listed in Table \ref{tab:human_eval}. The overall endorsement of raters for generated captions is 78.64\% for the full model and 62.42\% for the \textit{Pixel Only} model. These results indicate that our model can generate meaningful captions for UI elements. We found shorter captions tend to receive more rater endorsements than longer ones. The model with context still outperforms the one without context, which is consistent with automatic evaluation. See examples of captions generated by our model in Appendix~\ref{app:examples}.

\begin{table}[h!]
\small
\centering
\begin{tabular}{l|c|c|c}
    {Model} & {1+} & {2+} & {3+} \\
    \hline
   Pixel Only & 81.9 & 61.7 & 43.6 \\
   Pixel+Local+Context & \textbf{93.9} & \textbf{81.1} & \textbf{61.0} \\
\end{tabular}
  \caption{The human evaluation results. N+ in the header refers to N or more raters judge that the caption correctly describes the element.}
  \label{tab:human_eval}
\end{table}
\vspace{-5pt}
\subsection{Error Analysis}

To identify opportunities for improvements, we conducted error analysis on 50 widgets sampled from the validation set whose captions generated by the model share no words with their references. We classify these errors into the following types.

\begin{itemize}[nosep]
\item \textit{Nearby Elements} (21): The model is confused by nearby elements on the screen, e.g., outputting ``enter phone number" for ``write street address" on a sign-up screen.
\item \textit{Similar Appearance} (10): The model is confused by elements with a similar appearance, e.g., predicting ``delete" for an X-shaped image that is labeled as ``close".
\item \textit{Too Generic} (9): the model generate captions that are too generic, e.g., ``toggle on" instead of ``flight search on/off".
\item \textit{Model Correct} (10): The model produces semantically correct captions but treated as errors due to the limitation of automatic evaluation, e.g., ``close" for ``exit".
\end{itemize}

\noindent There are two directions for future improvement. One is to improve encoders for UI images and view hierarchies to better represent UI elements. The other is to improve data sparsity, which we want to better address long-tail phrases by expanding the dataset and having more elements and screens labeled.

\section{Conclusion}
We present widget captioning, a novel task for automatically generating language description for UI elements. The task is important because missing captions is a major issue for mobile accessibility and addressing the issue can improve accessibility and empower language-based mobile interaction in general. We created a large-scale dataset for this novel task by asking human annotators to create widget captions for a mobile UI corpus via crowdsourcing. We formulate widget captioning as a multimodal captioning task where both structural and image input are available. We experimented with a set of models based on the dataset. The winner configuration---a Transformer structural encoder coupled with a ResNet CNN---can generate semantically meaningful captions for sparsely labeled elements on the screen, which shows the feasibility of this task and opportunities for future research.

\section*{Acknowledgements}
We would like to thank Jason Baldridge for his valuable feedback, and our anonymous reviewers for their insightful comments that improved the paper.

\bibliography{anthology,emnlp2020}

\begin{thebibliography}{35}
\expandafter\ifx\csname natexlab\endcsname\relax\def\natexlab#1{#1}\fi

\bibitem[{AccessibilityScanner(2019)}]{apple_scanner}
AccessibilityScanner. 2019.
\newblock Apple accessibility scanner.
\newblock
  \url{https://developer.apple.com/library/archive/documentation/Accessibility/Conceptual/AccessibilityMacOSX/OSXAXTestingApps.html}.

\bibitem[{Anderson et~al.(2016)Anderson, Fernando, Johnson, and
  Gould}]{DBLP:journals/corr/AndersonFJG16}
Peter Anderson, Basura Fernando, Mark Johnson, and Stephen Gould. 2016.
\newblock \href {http://arxiv.org/abs/1607.08822} {{SPICE:} semantic
  propositional image caption evaluation}.
\newblock \emph{CoRR}, abs/1607.08822.

\bibitem[{Android(2019{\natexlab{a}})}]{android_classes}
Android. 2019{\natexlab{a}}.
\newblock Android widgets.
\newblock
  \url{https://developer.android.com/reference/kotlin/android/widget/package-summary}.

\bibitem[{Android(2019{\natexlab{b}})}]{android_support_classes}
Android. 2019{\natexlab{b}}.
\newblock android.support.v4.app.
\newblock
  \url{https://developer.android.com/reference/android/support/v4/app/package-summary}.

\bibitem[{AndroidLint(2019)}]{google_lint}
AndroidLint. 2019.
\newblock Improve your code with lint.
\newblock Https://developer.android.com/studio/write/lint.html.

\bibitem[{Chen et~al.(2015)Chen, Fang, Lin, Vedantam, Gupta, Dollar, and
  Lawrence~Zitnick}]{Chen2015-cy}
Xinlei Chen, Hao Fang, Tsung-Yi Lin, Ramakrishna Vedantam, Saurabh Gupta, Piotr
  Dollar, and C~Lawrence~Zitnick. 2015.
\newblock \href {http://arxiv.org/abs/1504.00325} {Microsoft {COCO} captions:
  Data collection and evaluation server}.

\bibitem[{Choo et~al.(2019)Choo, Balan, and
  Lee}]{Choo:2019:EAV:3290605.3300605}
Kenny Tsu~Wei Choo, Rajesh~Krishna Balan, and Youngki Lee. 2019.
\newblock \href {https://doi.org/10.1145/3290605.3300605} {Examining augmented
  virtuality impairment simulation for mobile app accessibility design}.
\newblock In \emph{Proceedings of the 2019 CHI Conference on Human Factors in
  Computing Systems}, CHI '19, pages 375:1--375:11, New York, NY, USA. ACM.

\bibitem[{Deka et~al.(2017)Deka, Huang, Franzen, Hibschman, Afergan, Li,
  Nichols, and Kumar}]{Deka:2017:RMA:3126594.3126651}
Biplab Deka, Zifeng Huang, Chad Franzen, Joshua Hibschman, Daniel Afergan, Yang
  Li, Jeffrey Nichols, and Ranjitha Kumar. 2017.
\newblock \href {https://doi.org/10.1145/3126594.3126651} {Rico: A mobile app
  dataset for building data-driven design applications}.
\newblock In \emph{Proceedings of the 30th Annual ACM Symposium on User
  Interface Software and Technology}, UIST '17, pages 845--854, New York, NY,
  USA. ACM.

\bibitem[{Denkowski and Lavie(2014)}]{Denkowski2014-mu}
Michael Denkowski and Alon Lavie. 2014.
\newblock Meteor universal: Language specific translation evaluation for any
  target language.
\newblock In \emph{Proceedings of the Ninth Workshop on Statistical Machine
  Translation}, pages 376--380, Stroudsburg, PA, USA. Association for
  Computational Linguistics.

\bibitem[{Devlin et~al.(2019)Devlin, Chang, Lee, and
  Toutanova}]{devlin-etal-2019-bert}
Jacob Devlin, Ming-Wei Chang, Kenton Lee, and Kristina Toutanova. 2019.
\newblock \href {https://doi.org/10.18653/v1/N19-1423} {{BERT}: Pre-training of
  deep bidirectional transformers for language understanding}.
\newblock In \emph{Proceedings of the 2019 Conference of the North {A}merican
  Chapter of the Association for Computational Linguistics: Human Language
  Technologies, Volume 1 (Long and Short Papers)}, pages 4171--4186,
  Minneapolis, Minnesota. Association for Computational Linguistics.

\bibitem[{{Donahue} et~al.(2017){Donahue}, {Hendricks}, {Rohrbach},
  {Venugopalan}, {Guadarrama}, {Saenko}, and {Darrell}}]{7558228}
J.~{Donahue}, L.~A. {Hendricks}, M.~{Rohrbach}, S.~{Venugopalan},
  S.~{Guadarrama}, K.~{Saenko}, and T.~{Darrell}. 2017.
\newblock Long-term recurrent convolutional networks for visual recognition and
  description.
\newblock \emph{IEEE Transactions on Pattern Analysis and Machine
  Intelligence}, 39(4):677--691.

\bibitem[{Gleason et~al.(2019)Gleason, Carrington, Cassidy, Morris, Kitani, and
  Bigham}]{Gleason:2019:LAL:3308558.3313605}
Cole Gleason, Patrick Carrington, Cameron Cassidy, Meredith~Ringel Morris,
  Kris~M. Kitani, and Jeffrey~P. Bigham. 2019.
\newblock \href {https://doi.org/10.1145/3308558.3313605} {\&ldquo;it's almost
  like they're trying to hide it\&rdquo;: How user-provided image descriptions
  have failed to make twitter accessible}.
\newblock In \emph{The World Wide Web Conference}, WWW '19, pages 549--559, New
  York, NY, USA. ACM.

\bibitem[{Guinness et~al.(2018)Guinness, Cutrell, and
  Morris}]{Guinness:2018:CCE:3173574.3174092}
Darren Guinness, Edward Cutrell, and Meredith~Ringel Morris. 2018.
\newblock \href {https://doi.org/10.1145/3173574.3174092} {Caption crawler:
  Enabling reusable alternative text descriptions using reverse image search}.
\newblock In \emph{Proceedings of the 2018 CHI Conference on Human Factors in
  Computing Systems}, CHI '18, pages 518:1--518:11, New York, NY, USA. ACM.

\bibitem[{He et~al.(2015)He, Zhang, Ren, and Sun}]{he2015residual}
Kaiming He, Xiangyu Zhang, Shaoqing Ren, and Jian Sun. 2015.
\newblock \href {http://arxiv.org/abs/1512.03385} {Deep residual learning for
  image recognition}.
\newblock Cite arxiv:1512.03385Comment: Tech report.

\bibitem[{Hodosh et~al.(2013)Hodosh, Young, and
  Hockenmaier}]{10.5555/2566972.2566993}
Micah Hodosh, Peter Young, and Julia Hockenmaier. 2013.
\newblock Framing image description as a ranking task: Data, models and
  evaluation metrics.
\newblock \emph{J. Artif. Int. Res.}, 47(1):853–899.

\bibitem[{Karpathy and Li(2014)}]{DBLP:journals/corr/KarpathyF14}
Andrej Karpathy and Fei{-}Fei Li. 2014.
\newblock \href {http://arxiv.org/abs/1412.2306} {Deep visual-semantic
  alignments for generating image descriptions}.
\newblock \emph{CoRR}, abs/1412.2306.

\bibitem[{Krizhevsky et~al.(2012)Krizhevsky, Sutskever, and
  Hinton}]{NIPS2012_4824}
Alex Krizhevsky, Ilya Sutskever, and Geoffrey~E Hinton. 2012.
\newblock \href
  {http://papers.nips.cc/paper/4824-imagenet-classification-with-deep-convolutional-neural-networks.pdf}
  {Imagenet classification with deep convolutional neural networks}.
\newblock In F.~Pereira, C.~J.~C. Burges, L.~Bottou, and K.~Q. Weinberger,
  editors, \emph{Advances in Neural Information Processing Systems 25}, pages
  1097--1105. Curran Associates, Inc.

\bibitem[{Li et~al.(2020)Li, He, Zhou, Zhang, and Baldridge}]{seq2act}
Yang Li, Jiacong He, Xin Zhou, Yuan Zhang, and Jason Baldridge. 2020.
\newblock \href {https://arxiv.org/abs/2005.03776} {Mapping natural language
  instructions to mobile ui action sequences}.
\newblock In \emph{Annual Conference of the Association for Computational
  Linguistics (ACL 2020)}.

\bibitem[{Lin and Och(2004)}]{Lin:2004:OME:1220355.1220427}
Chin-Yew Lin and Franz~Josef Och. 2004.
\newblock \href {https://doi.org/10.3115/1220355.1220427} {Orange: A method for
  evaluating automatic evaluation metrics for machine translation}.
\newblock In \emph{Proceedings of the 20th International Conference on
  Computational Linguistics}, COLING '04, Stroudsburg, PA, USA. Association for
  Computational Linguistics.

\bibitem[{Lin et~al.(2014)Lin, Maire, Belongie, Bourdev, Girshick, Hays,
  Perona, Ramanan, Doll{\'{a}}r, and Zitnick}]{DBLP:journals/corr/LinMBHPRDZ14}
Tsung{-}Yi Lin, Michael Maire, Serge~J. Belongie, Lubomir~D. Bourdev, Ross~B.
  Girshick, James Hays, Pietro Perona, Deva Ramanan, Piotr Doll{\'{a}}r, and
  C.~Lawrence Zitnick. 2014.
\newblock \href {http://arxiv.org/abs/1405.0312} {Microsoft {COCO:} common
  objects in context}.
\newblock \emph{CoRR}, abs/1405.0312.

\bibitem[{Niepert et~al.(2016)Niepert, Ahmed, and Kutzkov}]{pmlr-v48-niepert16}
Mathias Niepert, Mohamed Ahmed, and Konstantin Kutzkov. 2016.
\newblock \href {http://proceedings.mlr.press/v48/niepert16.html} {Learning
  convolutional neural networks for graphs}.
\newblock In \emph{Proceedings of The 33rd International Conference on Machine
  Learning}, volume~48 of \emph{Proceedings of Machine Learning Research},
  pages 2014--2023, New York, New York, USA. PMLR.

\bibitem[{Papineni et~al.(2002)Papineni, Roukos, Ward, and
  Zhu}]{Papineni2002-bw}
Kishore Papineni, Salim Roukos, Todd Ward, and Wei-Jing Zhu. 2002.
\newblock {BLEU}: a method for automatic evaluation of machine translation.
\newblock In \emph{Proceedings of the 40th Annual Meeting on Association for
  Computational Linguistics}, ACL '02, pages 311--318, USA. Association for
  Computational Linguistics.

\bibitem[{Pennington et~al.(2014)Pennington, Socher, and
  Manning}]{pennington2014glove}
Jeffrey Pennington, Richard Socher, and Christopher~D. Manning. 2014.
\newblock \href {http://www.aclweb.org/anthology/D14-1162} {Glove: Global
  vectors for word representation}.
\newblock In \emph{Empirical Methods in Natural Language Processing (EMNLP)},
  pages 1532--1543.

\bibitem[{Ross et~al.(2017)Ross, Zhang, Fogarty, and
  Wobbrock}]{Ross:2017:EFL:3132525.3132547}
Anne~Spencer Ross, Xiaoyi Zhang, James Fogarty, and Jacob~O. Wobbrock. 2017.
\newblock \href {https://doi.org/10.1145/3132525.3132547} {Epidemiology as a
  framework for large-scale mobile application accessibility assessment}.
\newblock In \emph{Proceedings of the 19th International ACM SIGACCESS
  Conference on Computers and Accessibility}, ASSETS '17, pages 2--11, New
  York, NY, USA. ACM.

\bibitem[{Ross et~al.(2018)Ross, Zhang, Fogarty, and
  Wobbrock}]{Ross:2018:EIB:3234695.3236364}
Anne~Spencer Ross, Xiaoyi Zhang, James Fogarty, and Jacob~O. Wobbrock. 2018.
\newblock \href {https://doi.org/10.1145/3234695.3236364} {Examining
  image-based button labeling for accessibility in android apps through
  large-scale analysis}.
\newblock In \emph{Proceedings of the 20th International ACM SIGACCESS
  Conference on Computers and Accessibility}, ASSETS '18, pages 119--130, New
  York, NY, USA. ACM.

\bibitem[{Sharma et~al.(2018)Sharma, Ding, Goodman, and
  Soricut}]{conceptual_captioning}
Piyush Sharma, Nan Ding, Sebastian Goodman, and Radu Soricut. 2018.
\newblock \href {https://doi.org/10.18653/v1/P18-1238} {Conceptual captions: A
  cleaned, hypernymed, image alt-text dataset for automatic image captioning}.
\newblock pages 2556--2565.

\bibitem[{Talkback(2019)}]{google_talkback}
Google~Android Talkback. 2019.
\newblock Android talkback.
\newblock
  \url{https://support.google.com/accessibility/android/answer/6007100?hl=en}.

\bibitem[{TensorFlow(2017)}]{tensorflow}
TensorFlow. 2017.
\newblock {An open-source software library for Machine Intelligence}.
\newblock \url{https://www.tensorflow.org/}.

\bibitem[{Vaswani et~al.(2017)Vaswani, Shazeer, Parmar, Uszkoreit, Jones,
  Gomez, Kaiser, and Polosukhin}]{NIPS2017_7181}
Ashish Vaswani, Noam Shazeer, Niki Parmar, Jakob Uszkoreit, Llion Jones,
  Aidan~N Gomez, \L~ukasz Kaiser, and Illia Polosukhin. 2017.
\newblock \href
  {http://papers.nips.cc/paper/7181-attention-is-all-you-need.pdf} {Attention
  is all you need}.
\newblock In I.~Guyon, U.~V. Luxburg, S.~Bengio, H.~Wallach, R.~Fergus,
  S.~Vishwanathan, and R.~Garnett, editors, \emph{Advances in Neural
  Information Processing Systems 30}, pages 5998--6008. Curran Associates, Inc.

\bibitem[{Vedantam et~al.(2015)Vedantam, Zitnick, and
  Parikh}]{DBLP:journals/corr/VedantamZP14a}
Ramakrishna Vedantam, C.~Zitnick, and Devi Parikh. 2015.
\newblock \href {https://doi.org/10.1109/CVPR.2015.7299087} {Cider:
  Consensus-based image description evaluation}.
\newblock pages 4566--4575.

\bibitem[{Vincent et~al.(2008)Vincent, Larochelle, Bengio, and
  Manzagol}]{Vincent:2008:ECR:1390156.1390294}
Pascal Vincent, Hugo Larochelle, Yoshua Bengio, and Pierre-Antoine Manzagol.
  2008.
\newblock \href {https://doi.org/10.1145/1390156.1390294} {Extracting and
  composing robust features with denoising autoencoders}.
\newblock In \emph{Proceedings of the 25th International Conference on Machine
  Learning}, ICML '08, pages 1096--1103, New York, NY, USA. ACM.

\bibitem[{Xu et~al.(2015)Xu, Ba, Kiros, Cho, Courville, Salakhutdinov, Zemel,
  and Bengio}]{DBLP:journals/corr/XuBKCCSZB15}
Kelvin Xu, Jimmy Ba, Ryan Kiros, Kyunghyun Cho, Aaron~C. Courville, Ruslan
  Salakhutdinov, Richard~S. Zemel, and Yoshua Bengio. 2015.
\newblock \href {http://arxiv.org/abs/1502.03044} {Show, attend and tell:
  Neural image caption generation with visual attention}.
\newblock \emph{CoRR}, abs/1502.03044.

\bibitem[{Young et~al.(2014)Young, Lai, Hodosh, and
  Hockenmaier}]{young-etal-2014-image}
Peter Young, Alice Lai, Micah Hodosh, and Julia Hockenmaier. 2014.
\newblock \href {https://doi.org/10.1162/tacl_a_00166} {From image descriptions
  to visual denotations: New similarity metrics for semantic inference over
  event descriptions}.
\newblock \emph{Transactions of the Association for Computational Linguistics},
  2:67--78.

\bibitem[{Zhang et~al.(2017)Zhang, Ross, Caspi, Fogarty, and
  Wobbrock}]{Zhang:2017:IPR:3025453.3025846}
Xiaoyi Zhang, Anne~Spencer Ross, Anat Caspi, James Fogarty, and Jacob~O.
  Wobbrock. 2017.
\newblock \href {https://doi.org/10.1145/3025453.3025846} {Interaction proxies
  for runtime repair and enhancement of mobile application accessibility}.
\newblock In \emph{Proceedings of the 2017 CHI Conference on Human Factors in
  Computing Systems}, CHI '17, pages 6024--6037, New York, NY, USA. ACM.

\bibitem[{Zhang et~al.(2018)Zhang, Ross, and
  Fogarty}]{Zhang:2018:RAM:3242587.3242616}
Xiaoyi Zhang, Anne~Spencer Ross, and James Fogarty. 2018.
\newblock \href {https://doi.org/10.1145/3242587.3242616} {Robust annotation of
  mobile application interfaces in methods for accessibility repair and
  enhancement}.
\newblock In \emph{Proceedings of the 31st Annual ACM Symposium on User
  Interface Software and Technology}, UIST '18, pages 609--621, New York, NY,
  USA. ACM.

\end{thebibliography}
\bibliographystyle{acl_natbib}

\appendix




\section{Further Details for Preprocessing the UI Corpus}
\label{app:proc_corpus}
We keep all the types of UI elements and determine the type of an element based on its class and ancestors attributes in the view hierarchy. We first check whether the element's class is in the set of predefined widget types in the Android development library~\cite{android_classes,android_support_classes}. If not, i.e., if it is a custom class, which is specific to an app, e.g., ``SearchButton", we find the closest class in its ancestry that belongs to the standard Android widget set as its type, e.g., ``Button". 

\section{The Annotation Interface}
\label{app:annotation_ui}

\begin{figure*}[h]
\centering
  \fbox{\includegraphics[width=1.4\columnwidth]{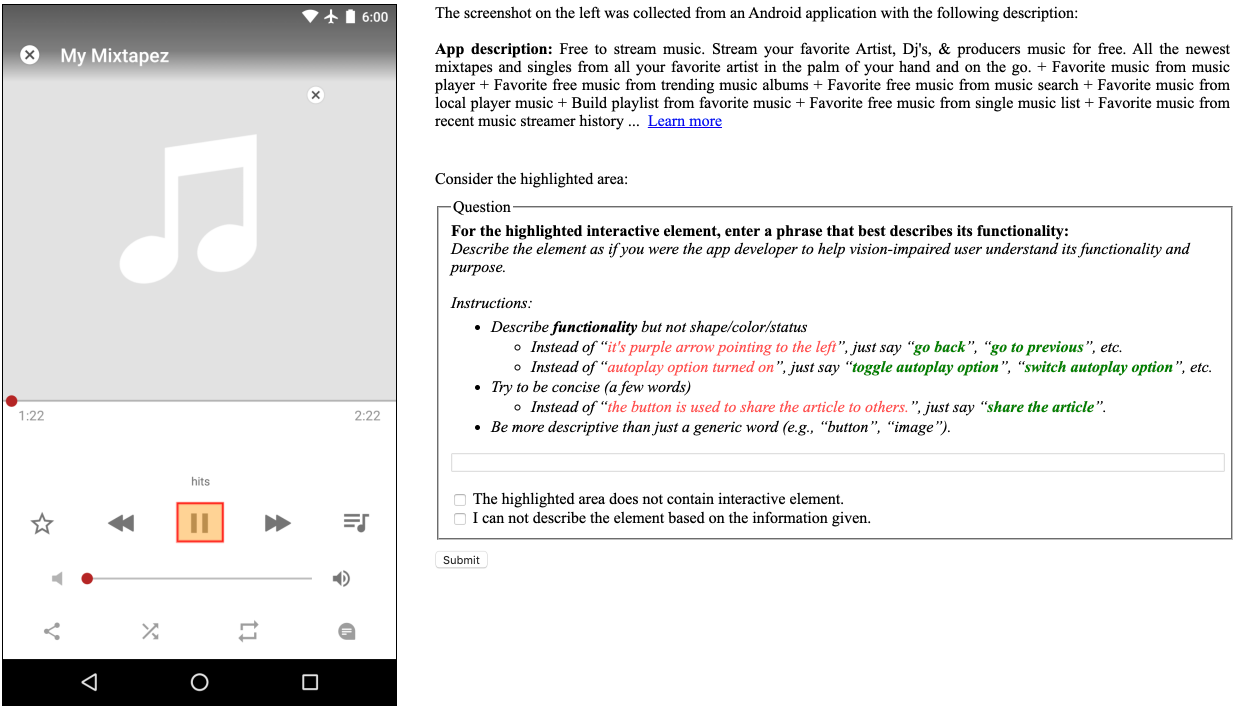}}
  \caption{The annotation web interface for crowd workers to create captions for a UI element.}~\label{fig:anontation}
  \vspace{-10pt}
\end{figure*}

We built a web interface for crowd workers to create captions for UI elements (see Figure~\ref{fig:anontation}). On the left of the interface is shown the screenshot of a mobile user interface. The element to be annotated is highlighted with a red bounding box. On the right, the app description that is crawled from Google Play Store is displayed to give the annotator the background about the mobile app that the UI screen is from. Underneath the app description the description for the annotation task. An annotator is given a guideline about the captioning task and several concrete examples about desired captions as well as captions should be avoided. The annotator can perform a task by entering a caption for the highlighted element in the text field, or skip the task by selecting the checkbox options that apply when the target element is incorrectly highlighted or cannot be described. The annotator clicks on the Submit button at the bottom to submit the responses.

\section{Phrase Distribution}
\label{app:phrase_dist}
\begin{figure*}[!htbp]
\centering
  \includegraphics[width=0.98\columnwidth]{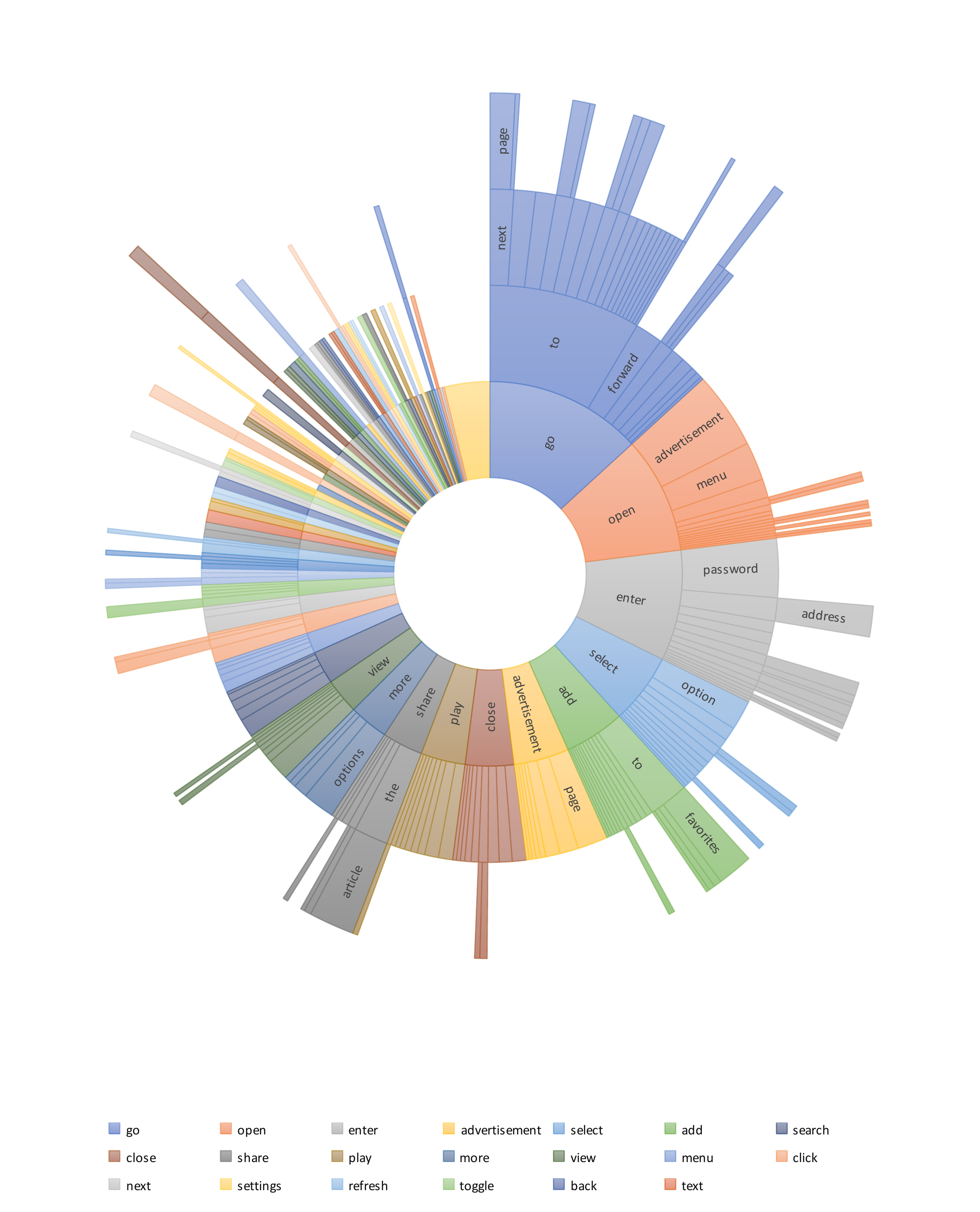}
  \caption{The distribution of captions by their first four words for the top 300 unique captions. The ordering of the words in a caption starts from the center and radiates outwards. The arc length is proportional to the number of captions containing the word. 
  }~\label{fig:caption_dist}
  \vspace{-20pt}
\end{figure*}

There are 476,912 UI elements in the UI corpus that come with text, which are from the app content or created by the app developers. These constitute the \texttt{widget\_text} feature input to the structural encoder to help generating descriptions for elements with missing captions. We compare the lengths of captions created by human workers with the lengths of these preexisting text content. We found the length distributions of the two sources are similar  (see Figure~\ref{fig:length_distribution}). The median length for both sources of text content is 2. There is a larger variance in length for the preexisting text, and there are more single-word and long descriptions. A pre-existing caption can simply repeat the content of an element that can be a long sentence or paragraph, which contributes to the long tail of the distribution (length$\geq{10}$). A pre-existing caption can also be generic names such as “image” or “button”, which are undesirable for accessibility. There are a diverse set of captions created by human workers (see Figure~\ref{fig:caption_dist}).
\begin{figure}[h]
\centering
  \includegraphics[width=0.5\columnwidth]{figures/rater_label_len.pdf}\includegraphics[width=0.5\columnwidth]{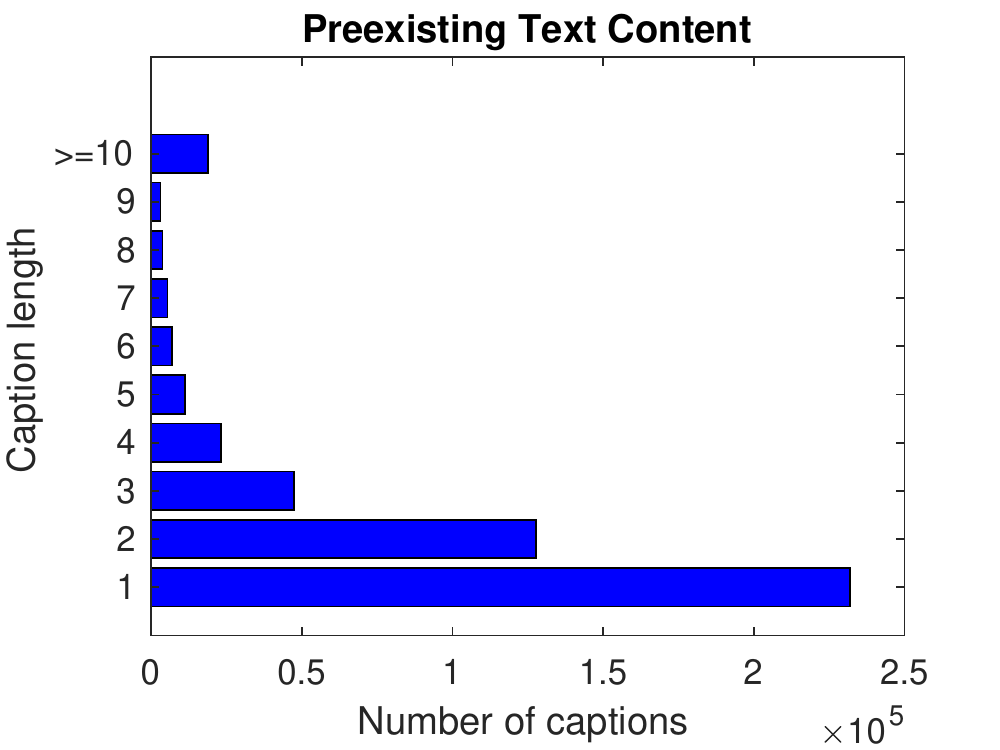}
  \caption{The length distribution of captions created by human workers versus those preexisting in the UI corpus. The X axis shows the number of text descriptions and the Y axis is the lengths from $1$ to $\geq{10}$.}~\label{fig:length_distribution}
\end{figure}



\section{Predicate-Object Phrases}
\label{app:predicate_obj_phrase}
We identify the \textit{Predicate + Object} subset used for evaluation in the paper as follows. First, we collected all the verbs with frequency more than 1000 in the corpus. This resulted in 22 verbs: \textit{go}, \textit{select}, \textit{enter}, \textit{open}, \textit{add}, \textit{search}, \textit{click}, \textit{toggle}, \textit{play}, \textit{view}, \textit{share}, \textit{close}, \textit{switch}, \textit{choose}, \textit{show}, \textit{download}, \textit{input}, \textit{see}, \textit{like}, \textit{change}, \textit{check}, and \textit{turn}. For these verbs, we manually checked all their objects and identified 194 nouns that are likely to require contextual and structural information from the view hierarchy to decode (see Table 1 in the paper). Finally, the \textit{Predicate + Object} captions were identified as the ones that contain at least one of the 22 verbs and one of the 194 nouns and appear at least twice in the corpus. As a result, 1850 (40.6\%) widgets in the test dataset have \textit{Predicate + Object} captions in their references and thus are selected as the \textit{Predicate + Object} subset for evaluating the model performance, as reported in the main paper.



\begin{table*}[h]
\small
\centering
\begin{tabularx}{0.82 \textwidth}{l|c|c|c|c|c|c}
    \hline
    {Model Configuration}
    & {BLEU-1}
    & {BLEU-2}
    & {ROUGE}
    & {CIDEr}
    & {METOER}
    & {SPICE} \\
    \hline
   & \multicolumn{6}{c}{Full Validation Set} \\
    \hline
    Template Matching & 19.5 & 10.1 & 20.1 & 35.2 & 12.6 & 5.8 \\
    Pixel Only & 35.8 & 23.9 & 35.9 & 70.7 & 24.6 & 10.8 \\
    Pixel+Local & 41.1 & 27.8 & 40.1 & 81.4 & 27.2 & 13.3 \\
    Pixel+Local+Context (PLC) & \textbf{44.6} & \textbf{30.2} & \textbf{43.9} & \textbf{91.8} & \textbf{29.9} & \textbf{16.3} \\
    PLC Classification & 36.2 & 24.4 & 36.7 & 76.3 & 25.4 & 12.9 \\
   \hline
   & \multicolumn{6}{c}{Predicate-Object Subset} \\
   \hline
    Template Matching & 19.0 & 8.9 & 19.2 & 27.0 & 10.9 & 6.4 \\
    Pixel Only & 38.2 & 24.4 & 38.0 & 64.0 & 23.8 & 13.0 \\
    Pixel+Local & 45.1 & 30.3 & 43.5 & 80.9 & 27.9 & 17.0 \\
    Pixel+Local+Context (PLC) & \textbf{51.0} & \textbf{35.1} & \textbf{49.7} & \textbf{100.3} & \textbf{32.9} & \textbf{21.7} \\
    PLC Classification & 38.9 & 26.6 & 38.9 & 76.9 & 26.0 & 16.7 \\
  \hline
\end{tabularx}
  \caption{The accuracy of each model configuration on the full set and the predicate-object subset of the validation dataset.}
  \label{tab:dev_result}
\end{table*}

\begin{table*}[h]
\small
\centering
\begin{tabularx}{0.71 \textwidth}{l|c|c|c|c|c|c}
    \hline
    {Ablation}
    & {BLEU-1}
    & {BLEU-2}
    & {ROUGE}
    & {CIDEr}
    & {METOER}
    & {SPICE} \\
    \hline
$-$ Text & 40.5 & 27.7 & 40.7 & 82.3 & 28.6 & 14.2 \\
$-$ Position & 43.4 & 30.8 & 43.6 & 93.8 & 30.8 & 16.2 \\
$-$ Widget Type & 44.3 & 30.9 & 43.6 & 92.1 & 30.3 & 16.4 \\
$-$ Clickable & 44.8 & 31.8 & 43.8 & 95.0 & 30.8 & 16.5 \\
$-$ Dom & 44.4 & 31.3 & 44.0 & 94.8 &30.9 & 16.9 \\
\hline
    Full model & \textbf{44.9} & \textbf{32.2} & \textbf{44.7} & \textbf{97.0} & \textbf{31.7} & \textbf{17.6} \\
   \hline
\end{tabularx}
  \caption{The ablation study results for the Pixel+Local+Context model.}
  \label{tab:ablation}
\end{table*}

\section{Model Configurations \& Training}
\label{app:config_train}
\textit{Template Matching} is based on a Nearest-Neighbor approach where all the examples in the training dataset are used as templates. Given an element to be captioned in the test dataset, the caption of the most similar template, based on cosine similarity between their pixel values, is used as the prediction.

For \textit{Pixel+Local}, the structural encoding is computed by feeding the element embedding, $e_i$, into a multi-layer perceptron that is followed by a linear projection: $\phi(e_{i}, \theta_{e})W_e$ where $\phi(\cdot)$ is a multi-layer perceptron parameterized by $\theta_{e}$, and $W_e$ are trainable parameters for the linear projection.

We pre-trained the ResNet image encoder using a denoised auto-encoder approach~\cite{Vincent:2008:ECR:1390156.1390294}. Pre-training allows us to leverage the images of all the elements instead of only those with caption labels. In particular, to reconstruct an image, we used 5 layers of transposed convolution where each layer has a residual connection architecture that is similar to the encoder part (that is discussed in the main paper). The reconstruction part of the model is discarded once the image encoder is trained. 

We implemented our model in TensorFlow~\cite{tensorflow}, and all the input and evalutation pipelines in Python. We tuned our models on a range of hyperparameters, including the hidden sizes (64, 128, 256 and 512), the number of encoder/decoder hidden layers (2, 4 and 6), widget text pooling (max, mean and sum), and the Transformer hyperparameter learning\_rate\_constant (0.01, 0.03, 0.1, 0.3, 1.0, 2.0).  We trained our model on 4 Tesla V100 GPU cores with asynchronous training with a batch size of 64 (screens) for all the models, which are all converged in less than 2 days. The model is trained, using the Adam optimizer, until it converges with a scheduled learning rate---a linear warmup followed by an exponential decay. We followed other modeling choices from the Transformer paper~\cite{NIPS2017_7181}. The number of parameters for \textit{Pixel Only}, \textit{Pixel+Local}, \textit{Pixel+Local+Context (PLC)} and \textit{PLC Classification} are 3.73M, 4.24M, 5.33M and 5.33M respectively. The performance of all the model configurations on the validation dataset are shown in Table \ref{tab:dev_result}.


\section{Ablation Study}
\label{app:ablation}
To investigate how different information in the view hierarchy contribute to the model performance, we conducted an ablation study by removing features in the UI element representation. As showed in Table \ref{tab:ablation}, each feature contributes to the overal performance of the model. Particularly, removing the text feature resulted in the largest drop in accuracy.

\section{Captioning Examples}
\label{app:examples}
\begin{figure}[!htbp]
\centering
\vskip 0pt
  \begin{subfigure}[t]{0.32\textwidth}
  \vskip 0pt
  \setlength{\fboxsep}{0pt}
  \fbox{\includegraphics[width=0.95\textwidth]{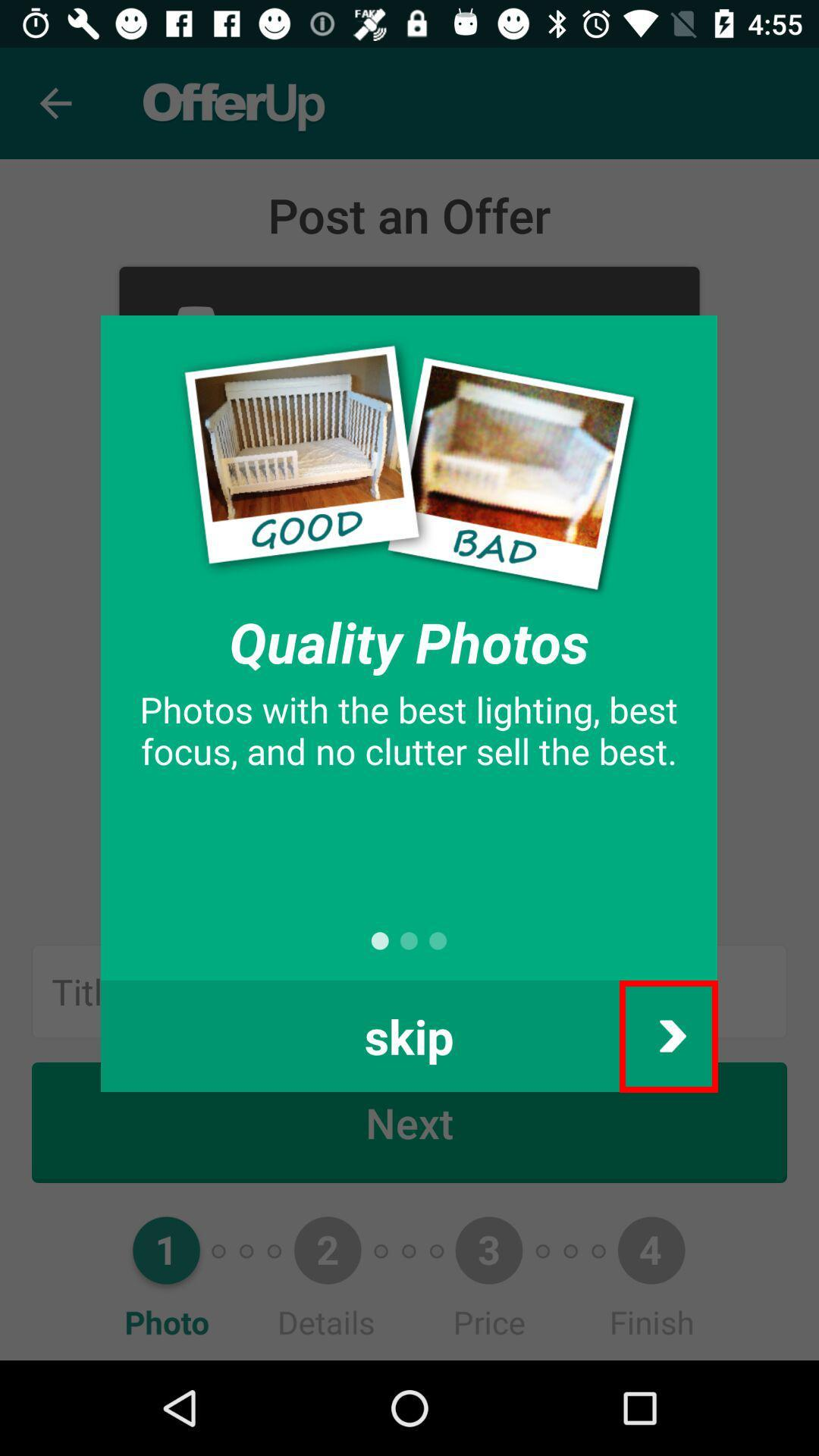}}
  \captionsetup{labelformat=empty}
  \caption{\textbf{Prediction}: go to next\\ \textbf{References}: next, skip button, toggle go to next option}
  \end{subfigure}%
  \hfill
  \begin{subfigure}[t]{0.32\textwidth}
  \vskip 0pt
  \setlength{\fboxsep}{0pt}
  \fbox{\includegraphics[width=0.95\textwidth]{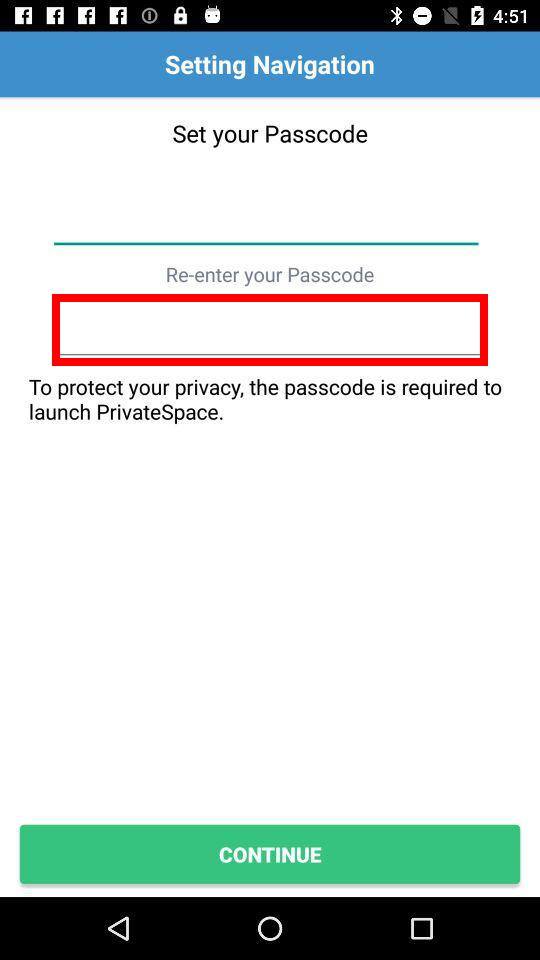}}
  \captionsetup{labelformat=empty}
  \caption{\textbf{Prediction}: enter password\\ \textbf{References}: text box, type your password}%
  \end{subfigure}%
  \hfill
  \begin{subfigure}[t]{0.32\textwidth}
  \vskip 0pt
  \setlength{\fboxsep}{0pt}
  \fbox{\includegraphics[width=0.95\textwidth]{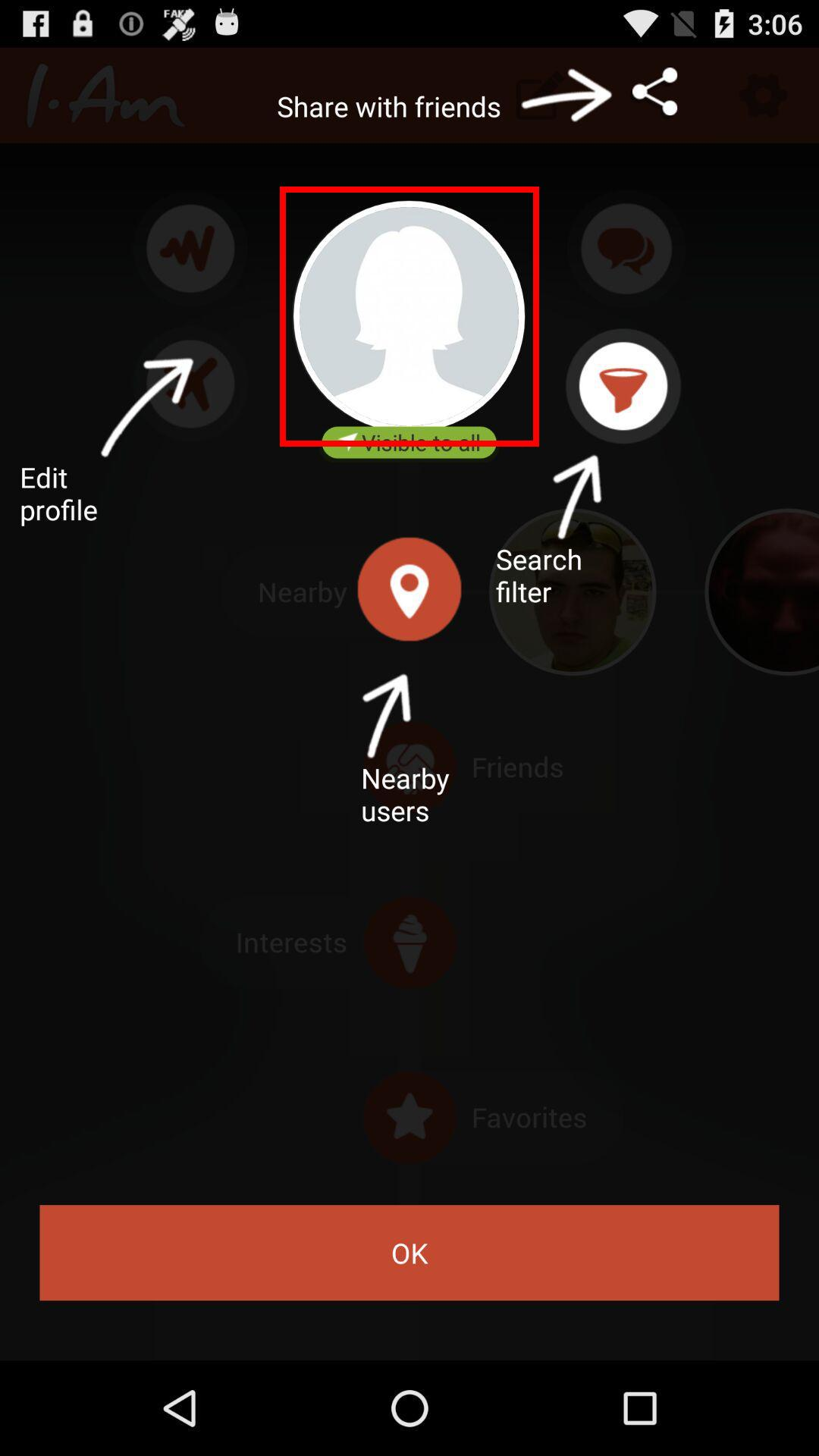}}
  \captionsetup{labelformat=empty}
  \caption{\textbf{Prediction}: profile picture\\ \textbf{References}: edit profile, edit profile photo, go to profile picture}
  \end{subfigure}%
  \\[5ex]
  \begin{subfigure}[t]{0.32\textwidth}
  \vskip 0pt
  \setlength{\fboxsep}{0pt}
  \fbox{\includegraphics[width=0.95\textwidth]{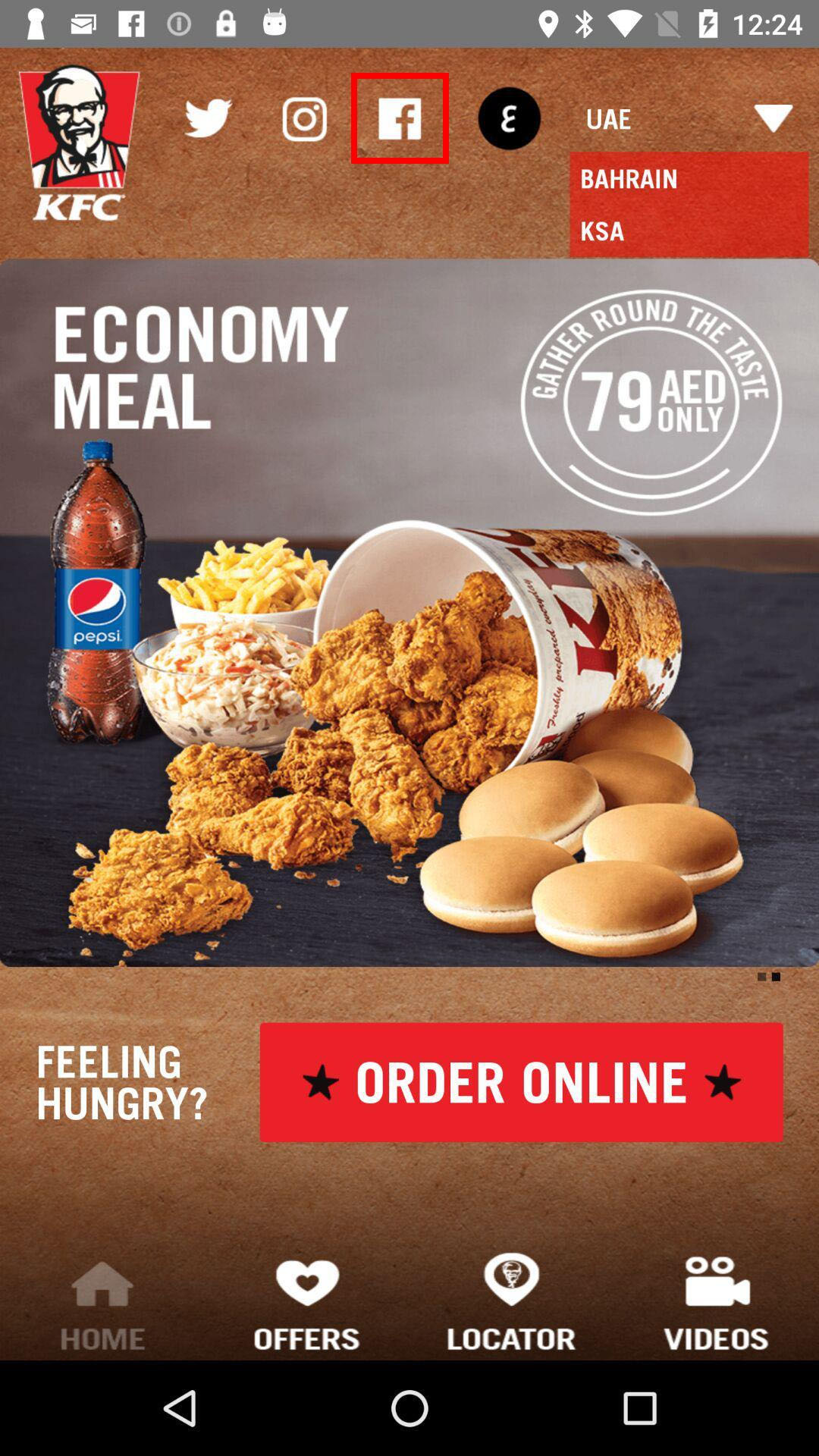}}
  \captionsetup{labelformat=empty}
  \caption{\textbf{Prediction}: share on facebook\\ \textbf{References}: facebook icon, facebook logo, see facebook page}
  \end{subfigure}%
  \hfill
  \begin{subfigure}[t]{0.32\textwidth}
  \vskip 0pt
  \setlength{\fboxsep}{0pt}
  \fbox{\includegraphics[width=0.95\textwidth]{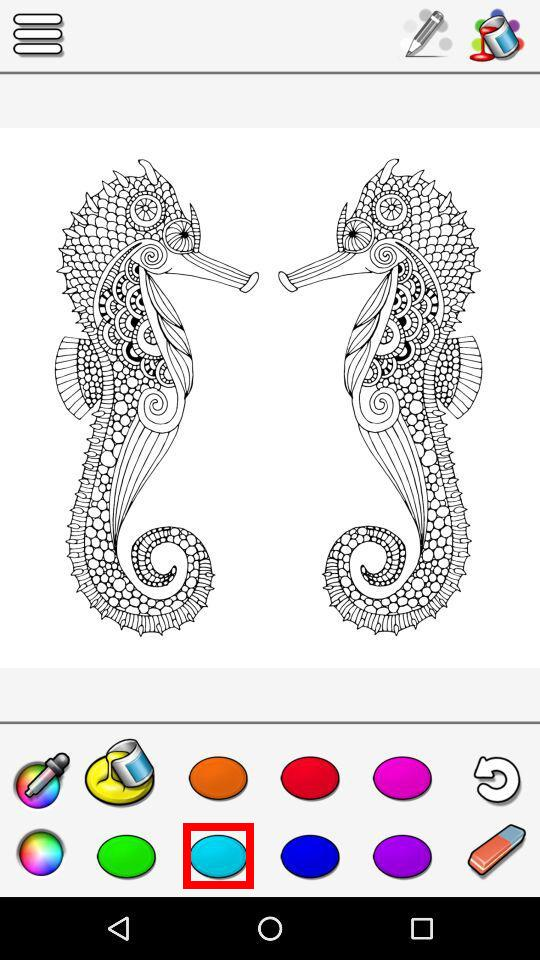}}
  \captionsetup{labelformat=empty}
  \caption{\textbf{Prediction}: select color\\ \textbf{References}: change color to teal, choose the color, pick a color}%
  \end{subfigure}%
  \hfill
  \begin{subfigure}[t]{0.33\textwidth}
  \vskip 0pt
  \setlength{\fboxsep}{0pt}
  \fbox{\includegraphics[width=0.95\textwidth]{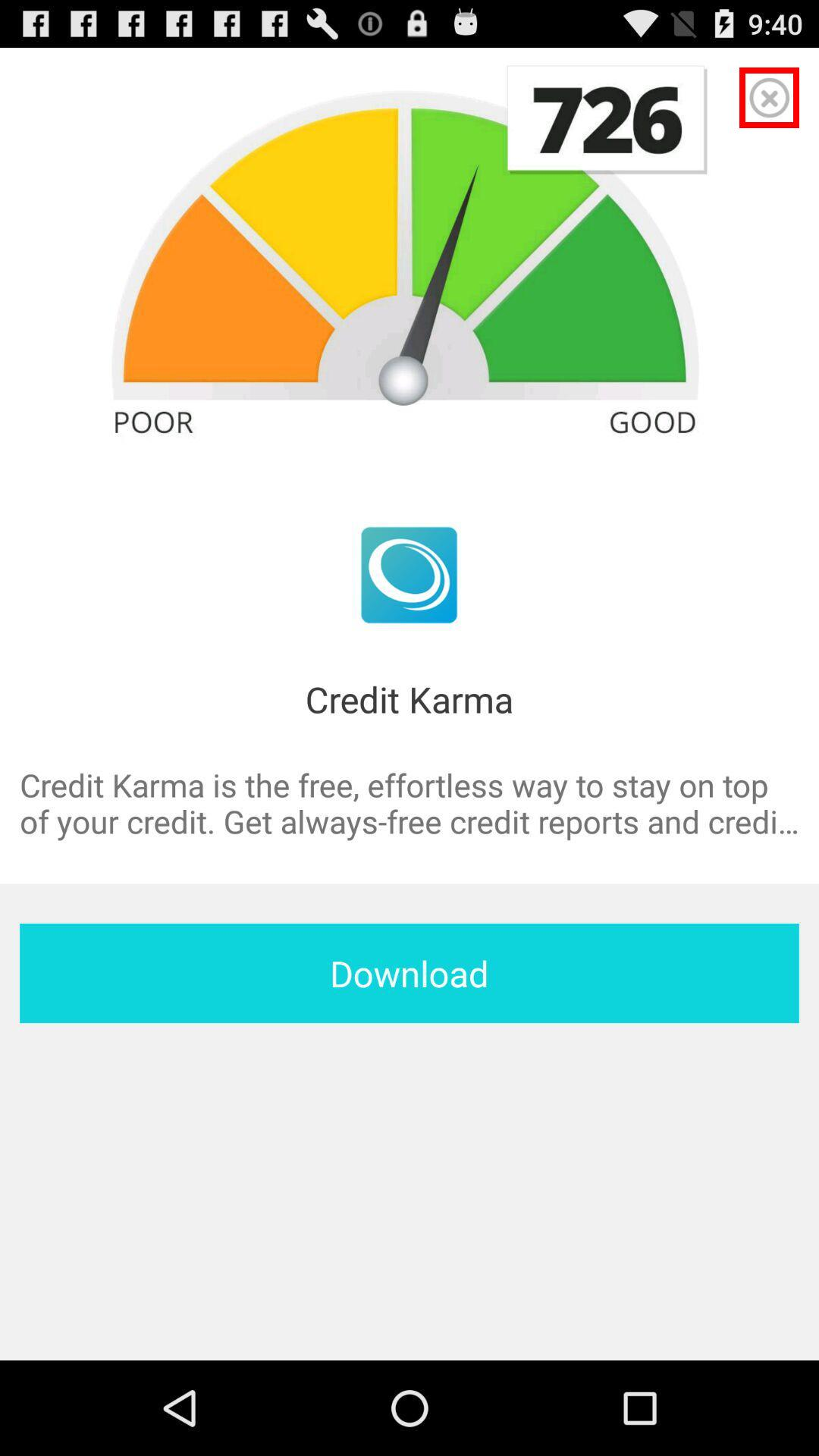}}
  \captionsetup{labelformat=empty}
  \caption{\textbf{Prediction}: close window\\ \textbf{References}: exit page, exit screen, exits out of current screen}
  \end{subfigure}%
  \caption{Widget captioning examples: the model predicted caption versus the reference captions labeled by human workers for the highlighted element in each screenshot.}~\label{fig:examples1}
\end{figure}

\begin{figure}[!htbp]
  \begin{subfigure}[t]{0.32\textwidth}
  \vskip 0pt
  \setlength{\fboxsep}{0pt}
  \fbox{\includegraphics[width=0.95\textwidth]{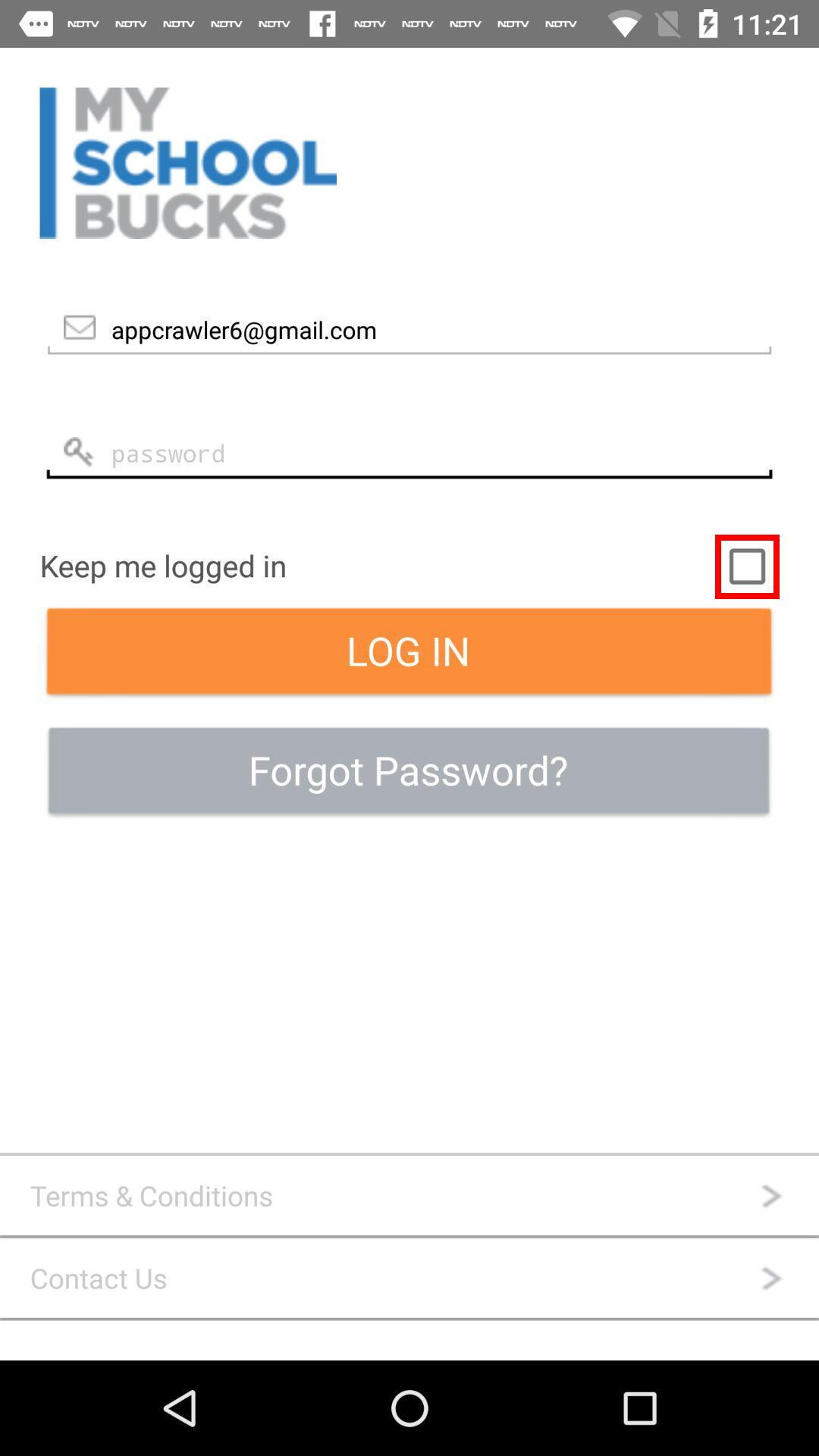}}
  \captionsetup{labelformat=empty}
  \caption{\textbf{Prediction}: toggle remember me option\\ \textbf{References}: check box for keep me logged in, check box to stay logged in, toggle check}
  \end{subfigure}%
  \hfill
  \begin{subfigure}[t]{0.32\textwidth}
  \vskip 0pt
  \setlength{\fboxsep}{0pt}
  \fbox{\includegraphics[width=0.95\textwidth]{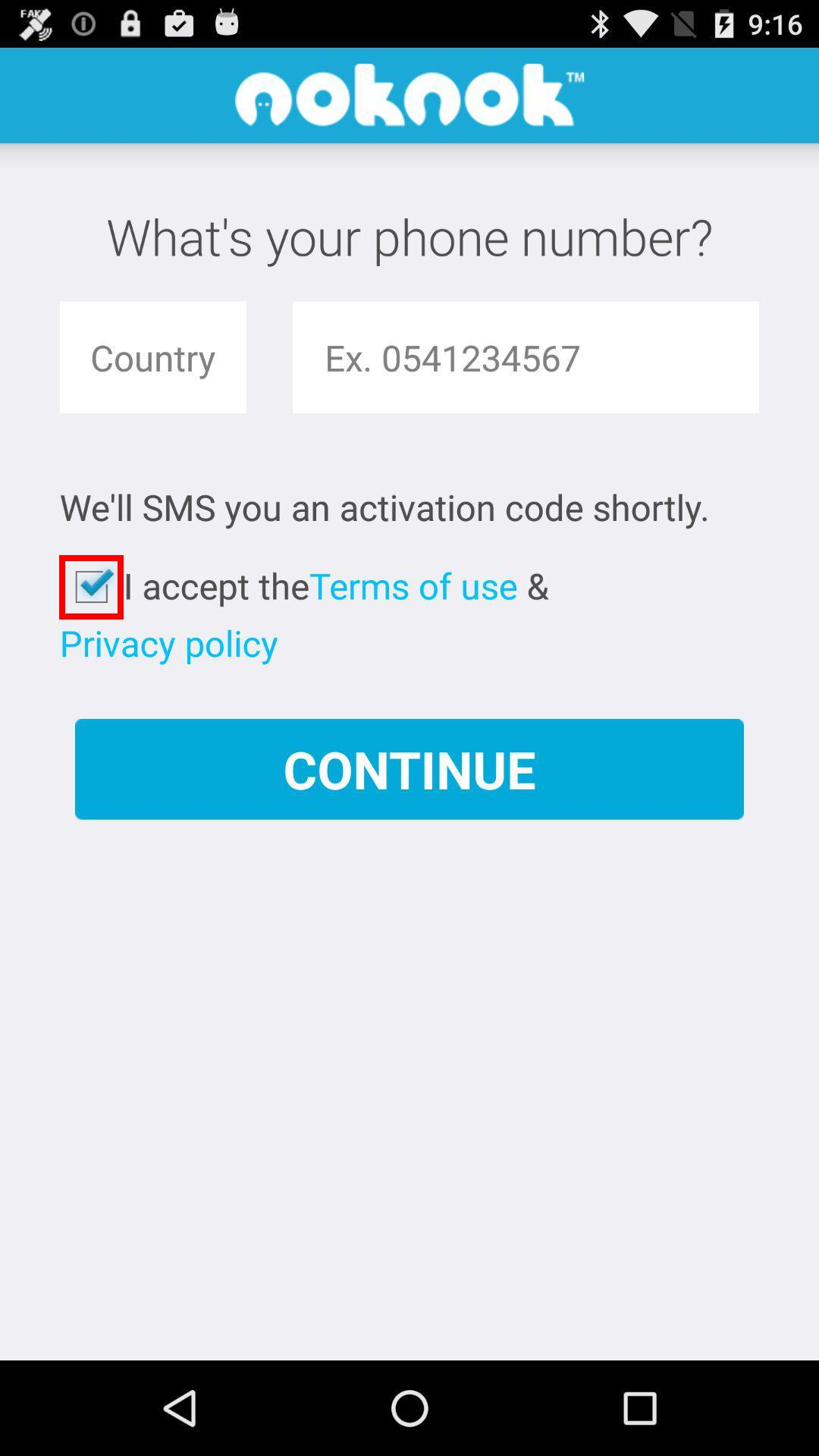}}
  \captionsetup{labelformat=empty}
  \caption{\textbf{Prediction}: check to agree to terms and conditions\\ \textbf{References}: accept, accept terms and conditions checkbox, toggle a select option}%
  \end{subfigure}%
  \hfill
\begin{subfigure}[t]{0.32\textwidth}
  \vskip 0pt
  \setlength{\fboxsep}{0pt}
  \fbox{\includegraphics[width=0.95\textwidth]{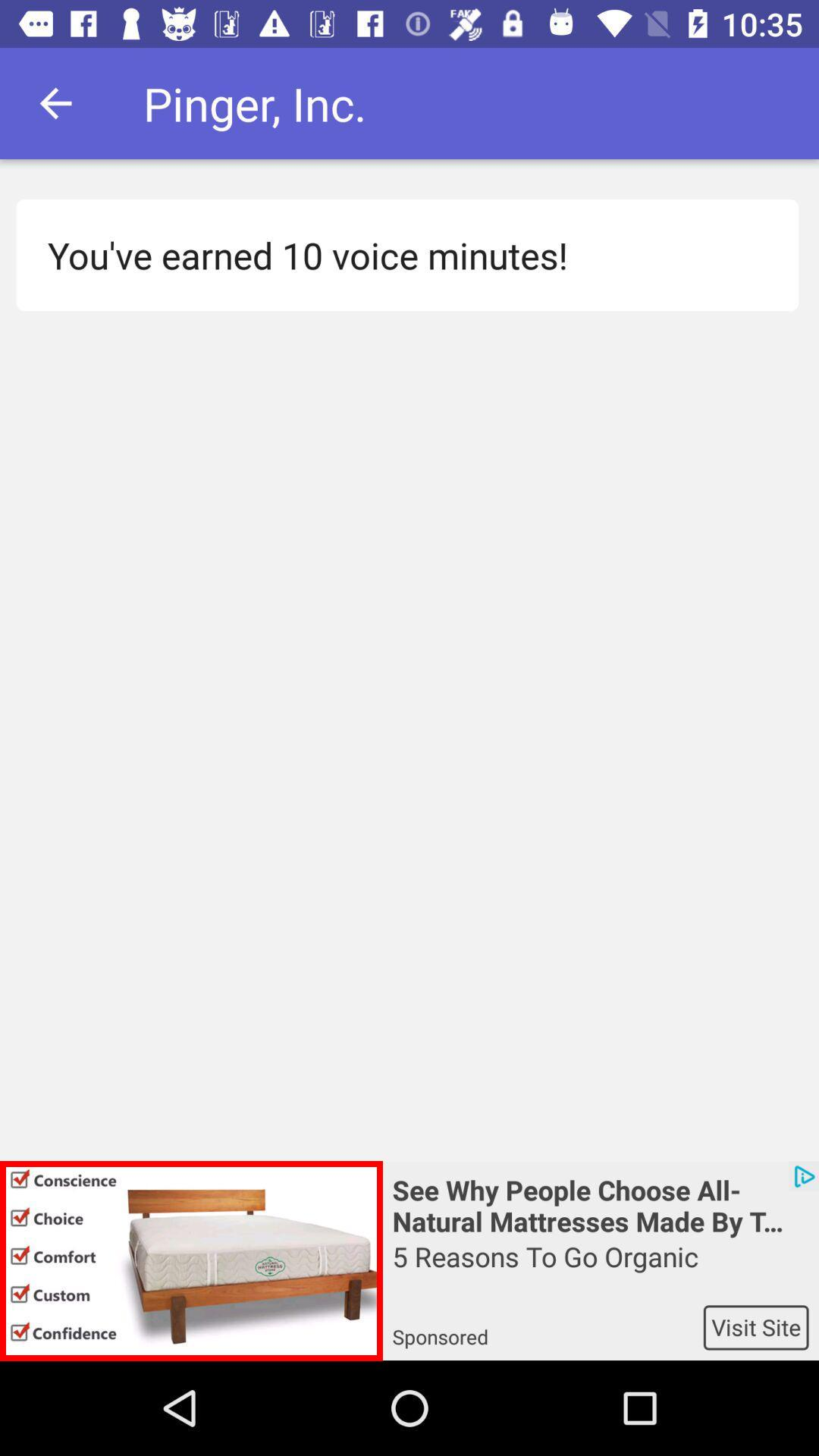}}
  \captionsetup{labelformat=empty}
  \caption{\textbf{Prediction}: go to advertisement\\ \textbf{References}: select advertisement, view advertisement}%
  \end{subfigure}%
  \\[5ex]
    \begin{subfigure}[t]{0.32\textwidth}
  \vskip 0pt
  \setlength{\fboxsep}{0pt}
  \fbox{\includegraphics[width=0.95\textwidth]{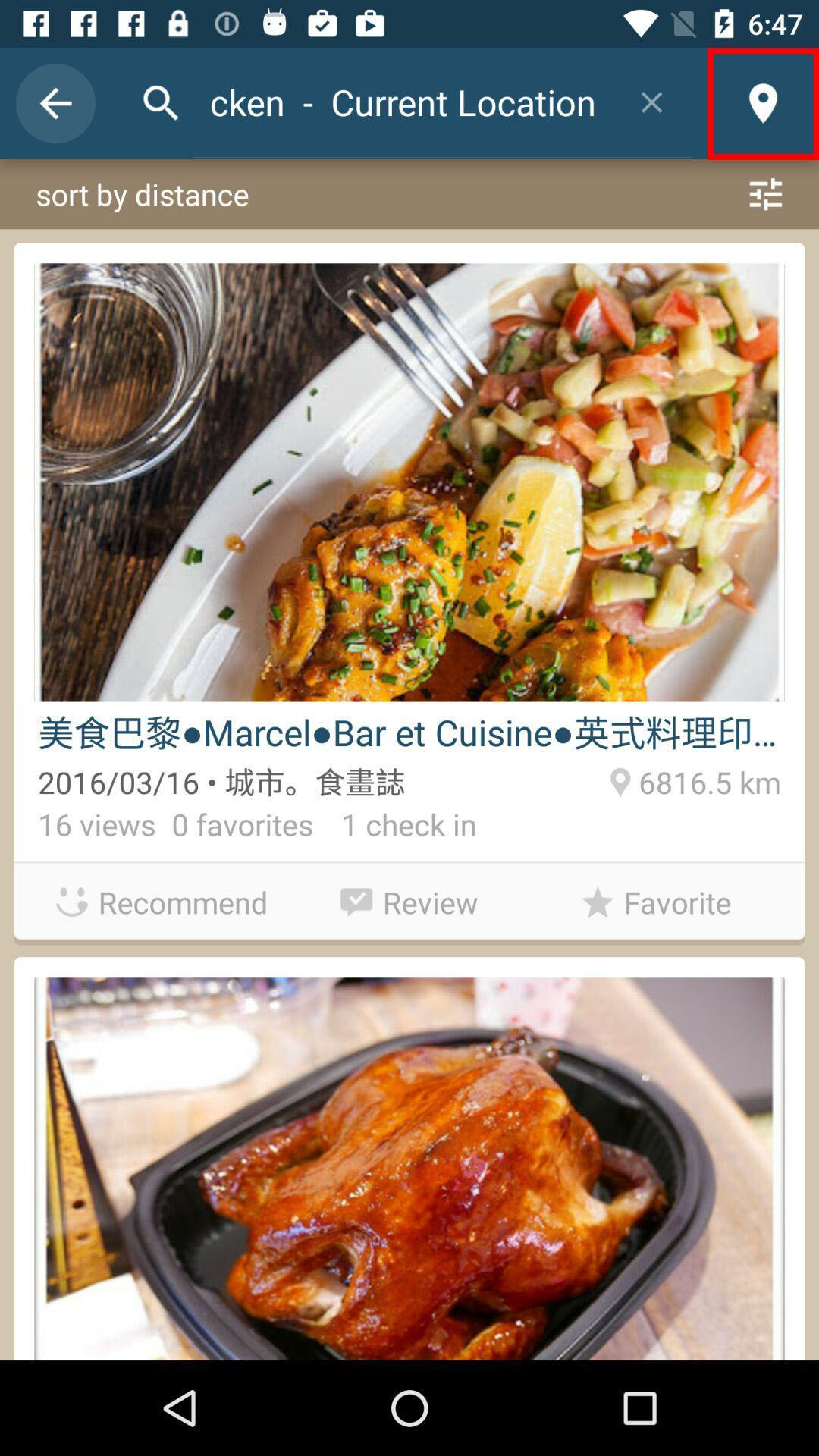}}
  \captionsetup{labelformat=empty}
  \caption{\textbf{Prediction}: go to location\\ \textbf{References}: choose location, open location settings, view map}%
  \end{subfigure}%
  \hfill
    \begin{subfigure}[t]{0.32\textwidth}
  \vskip 0pt
  \setlength{\fboxsep}{0pt}
  \fbox{\includegraphics[width=0.95\textwidth]{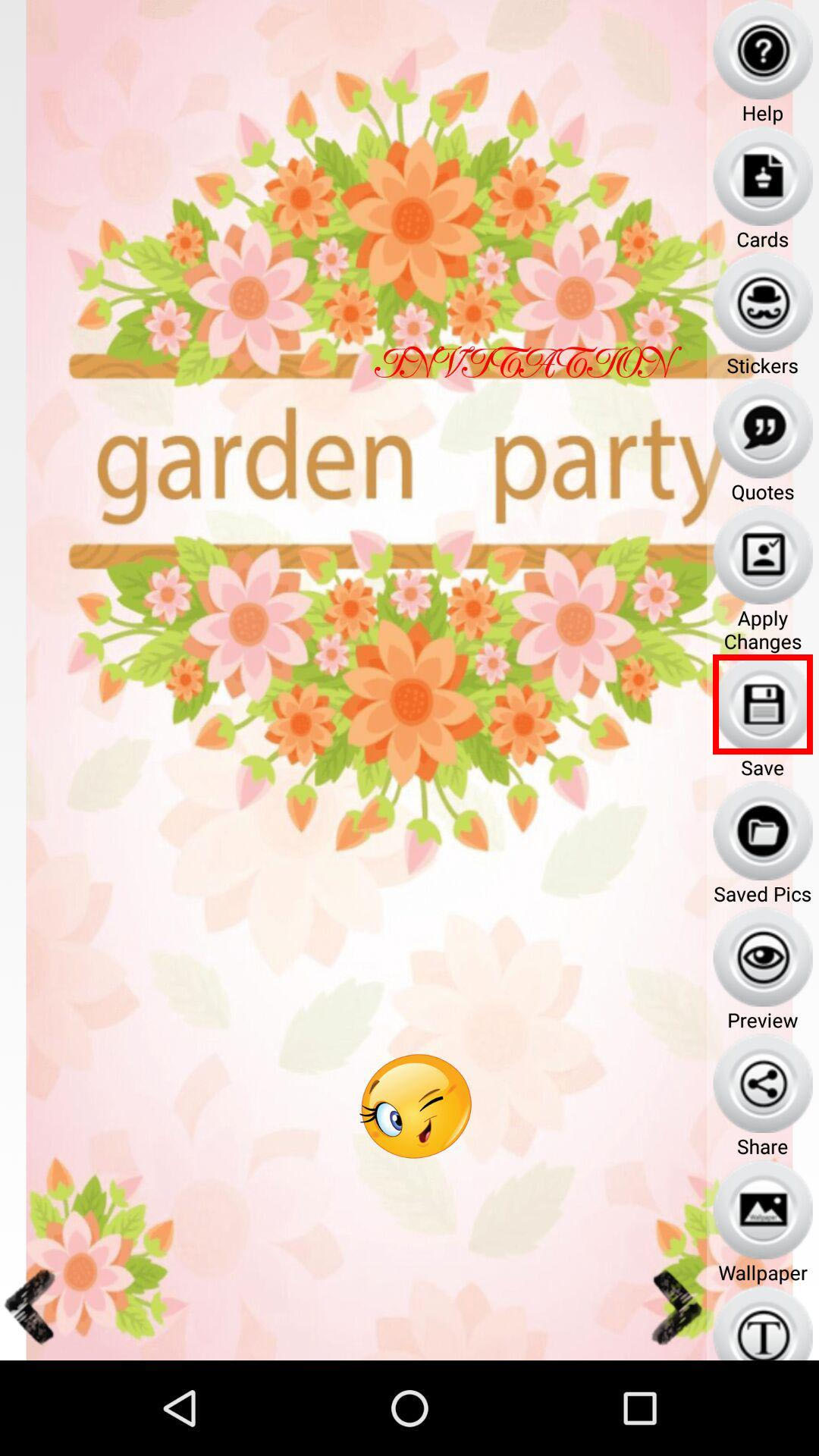}}
  \captionsetup{labelformat=empty}
  \caption{\textbf{Prediction}: save\\ \textbf{References}: save, save image, save template}%
  \end{subfigure}%
  \hfill
    \begin{subfigure}[t]{0.32\textwidth}
  \vskip 0pt
  \setlength{\fboxsep}{0pt}
  \fbox{\includegraphics[width=0.95\textwidth]{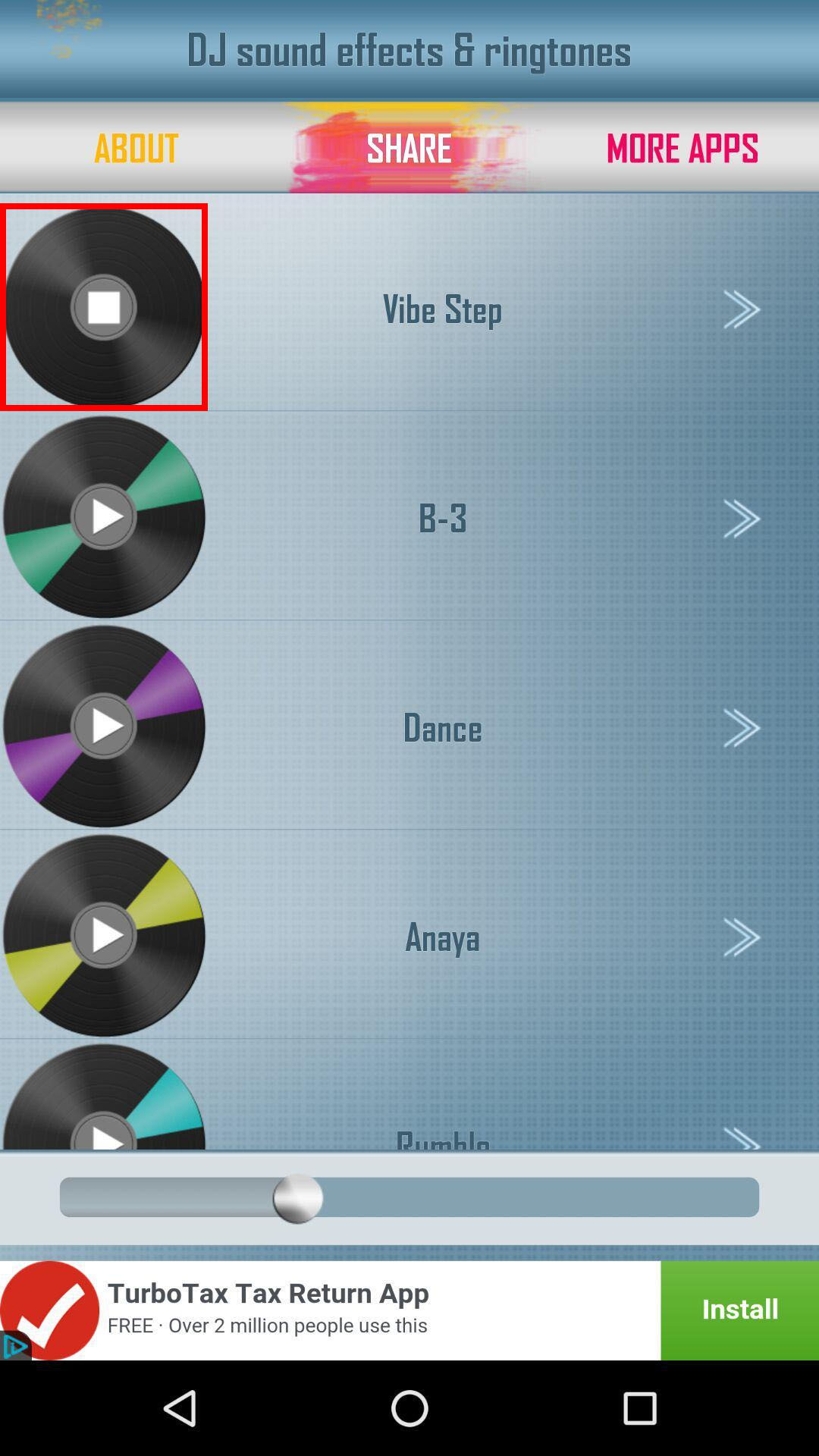}}
  \captionsetup{labelformat=empty}
  \caption{\textbf{Prediction}: play music\\ \textbf{References}: playing audio, stop sound}%
  \end{subfigure}%
  \hfill
  \label{fig:examnples}
\caption{More examples: the model predicted caption versus the reference captions labeled by human workers for the highlighted element in each screenshot.}~\label{fig:examples2}
\end{figure}

A collection of examples of captions generated by the models versus those created by human workers from the held-out test dataset are shown in Figure~\ref{fig:examples1} and~\ref{fig:examples2}.
\balance

\onecolumn

\end{document}